\documentclass[twocolumn]{galbot}
\usepackage{amsmath,amssymb}
\usepackage{array}
\usepackage{booktabs}
\usepackage{colortbl}
\usepackage{adjustbox}
\usepackage{xspace}
\usepackage[shortlabels]{enumitem}
\usepackage{placeins}

\usepackage{algorithm}
\usepackage{algpseudocode}
\crefname{algorithm}{Algorithm}{Algorithms}
\Crefname{algorithm}{Algorithm}{Algorithms}

\newcommand{\method}{Harness$^2$\xspace}

\makeatletter
\DeclareRobustCommand\onedot{\futurelet\@let@token\@onedot}
\def\@onedot{\ifx\@let@token.\else.\null\fi\xspace}
\def\eg{\emph{e.g}\onedot} 
\def\ie{\emph{i.e}\onedot}

\makeatother

\definecolor{hsOrange}{HTML}{C4673E}
\definecolor{hsOrangeBg}{HTML}{F9E9DF}
\definecolor{hsBlue}{HTML}{3F6F97}
\definecolor{hsBlueBg}{HTML}{E6EEF5}
\definecolor{hsPurple}{HTML}{6A58B0}
\definecolor{hsPurpleBg}{HTML}{ECE8F7}
\definecolor{hsGreen}{HTML}{3A8A79}
\definecolor{hsGreenBg}{HTML}{DFF0EA}
\definecolor{hsGrayBg}{HTML}{F4F6F8}
\newcommand{\tabstyle}{%
  \setlength{\aboverulesep}{0pt}\setlength{\belowrulesep}{0pt}%
  \renewcommand{\arraystretch}{1.18}%
  \arrayrulecolor{hsBlue}}
\newcommand{\headrow}{\rowcolor{hsBlueBg}}
\newcommand{\oursrow}{\rowcolor{hsOrangeBg}}
\newcommand{\altrow}{\rowcolor{hsGrayBg}}
\newcommand{\hd}[1]{\textbf{\color{hsBlue}#1}}
\newcommand{\grouphd}[2]{\textbf{\color{#1}#2}}

\makeatletter
\providecommand\@LN[2]{}
\providecommand\@LN@col[1]{}
\makeatother

\title{Harness$^2$: Building Robotic Harnesses\\ with a VLM Harness for Fine-Grained Manipulation}

\author[1,6,*]{Zekai~Li}
\author[2,6,*]{Jiayi~Yu}
\author[2,6,*]{Jinlu~Zhang}
\author[3,6,*]{Leyao~Wei}
\author[2,6,*]{Xinze~Liu}
\author[1,6,*]{Zekun~Qi}
\author[2,6]{Yuxuan~Wan}
\author[2,6]{Luyang~Niu}
\author[4,6]{Sikai~Liang}
\author[1,6]{Ruixi~Yu}
\author[3,6]{Xuchuan~Chen}
\author[1,6]{Zhuo~Cao}
\author[5,6]{Dairu~Liu}
\author[2,6]{Yintianrun~Zhang}
\author[1,6]{YingSheng~Zhang}
\author[2,6,\dagger]{He~Wang}
\author[2,\dagger]{Yizhou~Wang}
\author[1,6,\dagger]{Li~Yi}

\affiliation[1]{Tsinghua University}
\affiliation[2]{Peking University}
\affiliation[3]{Beihang University}
\affiliation[4]{Shanghai Jiao Tong University}
\affiliation[5]{Nankai University}
\affiliation[6]{Galbot}
\contribution[*]{Equal contribution}
\contribution[\dagger]{Corresponding authors}

\date{\today}
\page{\url{https://galaxygeneralrobotics.github.io/Harness-2/}}

\abstract{
Robot agents built on vision-language models (VLMs) now lead general manipulation benchmarks, yet they still fail at \textbf{fine-grained manipulation}, where success depends on millimeter alignment and stable contact.
These agents improve by rewriting prompts, programs, and skill libraries around fixed models and generic tools, but what fine-grained tasks lack cannot be written: perception precise to the millimeter, and control fast and reliable enough for contact.
These capabilities must be trained.
Yet for each new task, building the training pipeline for perception and control is still human engineering.
We present \method, in which a coding agent, guided by a single task-independent method note, automates this entire pipeline.
In simulation, the agent builds the harness that a VLM deploys: a visuomotor skill for each stage that the VLM cannot control, learned from reinforcement-learning experts, and a visual localizer for each stage that it cannot perceive precisely.
No task-specific prompting or human engineering is required.
On eight fine-grained RoboDojo tasks, \method raises average success to 69.0\% from 38.5\% for the best previous performance per task; on a real dual-arm robot, without real-robot demonstrations, it outperforms a VLA trained on 100 real demonstrations per task on both insertion tasks.
}

\begin{document}
\maketitle

\begin{figure*}[!t]
  \centering
  \includegraphics[width=0.9\textwidth]{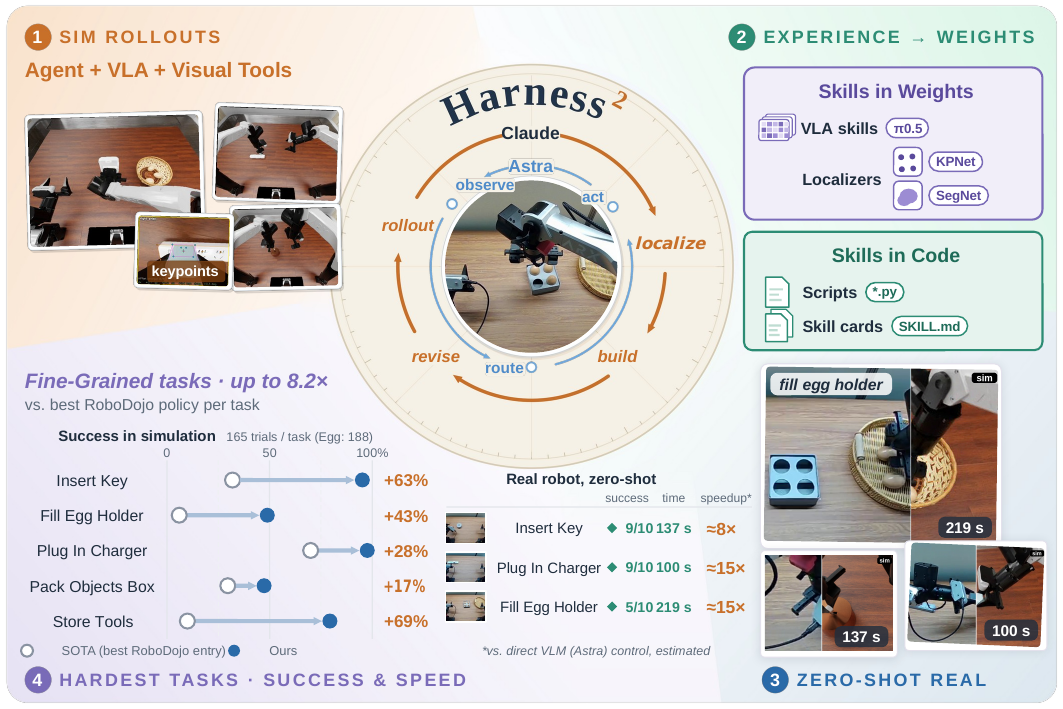}
  \caption{\textbf{Overview of \method.}
  A VLM used directly as the policy fails at fine-grained manipulation: some stages need geometry it cannot perceive at the millimeter level, and others need control more reliable and faster than its calls.
  In \method, an outer harness (Claude) rolls out the inner harness, the VLM (Astra) with its tools, in simulation (1) and builds the tools that the VLM lacks (2): skills in weights (VLA skills, visual localizers) and skills in code (scripts, skill cards).
  The frozen inner harness is then deployed zero-shot on the real robot (3).
  Against the best RoboDojo entry on each task, success in simulation rises by up to 8.2$\times$, and a real trial runs an estimated 8--15$\times$ faster than direct VLM control (4).}
  \label{fig:teaser}
\end{figure*}

% Restructured arXiv version (2026-11-30, v2d): situation -> how current agents improve -> what fine-grained tasks lack
% -> the question we ask -> what an agent must answer and how Harness2 answers it -> results -> contributions.
% Method details (takeover probe, inner/outer terminology, router) are left to Sec. 3.
\section{Introduction}
\label{sec:intro}

% P1: situation -- agents lead, but the tasks that remain unsolved are the fine-grained ones
Robot agents built on vision-language models (VLMs) now lead general manipulation benchmarks.
On RoboDojo~\cite{chen2026robodojo,community2026xpolicylab}, the best agent harness averages 68.9\% success over all 42 tasks at the end of October 2026, against 35.0\% for the best vision-language-action (VLA) model.
The tasks that remain unsolved, however, are the fine-grained ones, in which success depends on millimeter alignment, stable contact, and objects that must not slip in the grasp.
On the five tasks with the lowest best-entry score, such as placing eggs in a holder or storing tools in a toolbox, the best entry per task averages 22.8\%, and none reaches 50\%; on inserting a key, the best entry reaches 32.0\%~(\Cref{tab:main_sim}).

% P2: how current agents get better -- text and code around fixed models and generic tools
Current agents improve by rewriting what surrounds the model.
They control the robot by writing code over general perception and motion tools, typically promptable segmentation with multi-view geometry~\cite{gao2026hexashell,si2026pyrualean,wang2026auriga}, and they improve from experience by revising prompts, programs, skill libraries, and execution records while the weights of every model stay fixed~\cite{li2026physrsieval,chen2026recursia,yu2026robodoctor,sun2026reskill,guan2026seek}.
The alternative, fine-tuning a VLA on demonstrations of each task~\cite{black2025pi05,kim2025openvlaoft,nvidia2025gr00t}, does not reach these tasks either: $\pi_{0.5}$ fine-tuned on simulated demonstrations averages 0.8\% on the eight tasks we study.

% P3: what fine-grained tasks lack, and why text and code cannot supply it
What fine-grained tasks lack is not reasoning but two capabilities that text and code cannot supply.
The first is precise perception: these tasks require millimeter-level alignment, which the segmentation-based localization of current agents does not reliably reach.
The second is fast and reliable control: a VLM call takes about 25\,s, so contact-rich motion runs open loop, and the same VLM acts differently in repeated runs of the same scene.
Both capabilities must be trained: a localizer that measures the offset to be corrected, and a visuomotor skill that executes contact at the control rate.
Training them for each task is the engineering that agents were meant to remove, and today it is done by hand.

% P4: the question, and what an agent must answer to do it
Given a simulated task, can a frozen VLM train or write the tools that VLM lacks without task-specific human engineering?
An agent that trains tools must answer three questions that an agent writing code never faces.
\emph{What should be trained?} A task is a chain of stages, and different stages lack different capabilities; training a skill where perception is missing, or a localizer where control is missing, wastes hours of training.
\emph{Where do the training data come from?} Fine-grained skills need demonstrations of exactly the motions that are hardest to teleoperate precisely.
\emph{Where should the effort go?} Training and evaluation are expensive, and an episode succeeds only if every stage does.

% P5: Harness2 -- how it answers the three questions
We present \method, a harness that builds a harness~(\Cref{fig:teaser}).
An off-the-shelf coding agent, guided by a method note that gives a generic training recipe and is the same for every task, builds in simulation the harness that the VLM deploys, and answers each question with a measurement.
To decide what to train, it tests on each stage whether a policy that sees only camera images can do what an expert that sees the true simulator state does: if it can, the stage needs a visuomotor skill; if it cannot, the missing capability is perception, and the stage needs a visual localizer; stages that the VLM handles in the same way every time are compiled into scripts.
The training data come from these simulator experts, trained with reinforcement learning.
The effort goes to the first stage whose measured success falls short, until every stage passes.
At deployment, only the VLM and the built tools run, from RGB images and proprioception, in simulation and on the real robot.

% P6: results
On eight fine-grained RoboDojo tasks, evaluated on all official layouts, \method reaches 69.0\% average success against 38.5\% for the best leaderboard entry per task (\Cref{tab:main_sim}).
Deployed zero-shot on a real dual-arm robot, it outperforms $\pi_{0.5}$ post-trained on 100 real demonstrations per task on both insertion tasks (\Cref{tab:real}).
Starting from an empty experience library, the agent builds harnesses that exceed the best leaderboard entry within hours (\Cref{sec:exp_build}).

\noindent Our contributions are:
\begin{itemize}[leftmargin=*,itemsep=1pt,topsep=2pt]
  \item \textbf{From writing tools to training them.} We show that what VLM agents lack on fine-grained manipulation, precise perception and reliable contact control, cannot be supplied by text and code and must be trained, and we pose the problem of having the agent itself train it.
  \item \textbf{\method.} A coding agent that builds a VLM's harness, including its learned tools, from a task-independent method note: it decides per stage which tool to train by a measurement in simulation, trains it from simulator experts, and repairs the first failing stage (\Cref{sec:exp_diag,sec:exp_build}).
  \item \textbf{Fine-grained manipulation in simulation and on a real robot.} The best results on eight fine-grained RoboDojo tasks as of 2 November 2026; zero-shot real-robot deployment, which outperforms a VLA trained on 100 real demonstrations per task on both insertion tasks; and harnesses built from scratch that exceed the best leaderboard entries within hours (\Cref{sec:exp_main,sec:exp_build,sec:exp_eff}).
\end{itemize}

\section{Related Work}
\label{sec:related}

\paragraph{VLM agents and harnesses for manipulation.}
Early planners use a language model to call perception and motion primitives~\cite{huang2023voxposer,huang2024rekep,liang2023cap}.
More recent systems add memory and skill libraries around a VLM to handle long-horizon tasks~\cite{li2026roboclaw}, geometric reasoning~\cite{wang2026auriga}, or VLA steering~\cite{zhang2026harnessvla}, and the harness design itself can be optimized from experience~\cite{lee2026emho}.
Others use the VLM directly as a policy~\cite{openai2026astra}.
VLAs-as-Tools~\cite{lei2026vlastools} lets a VLM dispatch task-specific VLA skills; CaP-X~\cite{fu2026capx} benchmarks coding agents for robot control; Robo-Harness K1~\cite{chen2026roboharnessk1} adds calibrated perception tools.
In all of these, the tools are hand-written, generated as code, or fine-tuned VLAs with a fixed interface.
\method instead \emph{trains} both controllers and localizers, chosen by what each stage lacks, and replaces routine VLM behavior with scripts where appropriate.

\paragraph{Agents that automate robot learning.}
Language models write rewards and curricula~\cite{lee2026rda,ma2024eureka,ma2024dreureka}, generate scenes, tasks, and demonstrations~\cite{wang2024gensim,wang2024robogen}, and grow libraries of control programs by trial and repair~\cite{lu2026aspire}.
The closest work closes a physical feedback loop around a training pipeline, edits the training algorithm, and transfers experience across tasks~\cite{xiao2026enpire}, or separates the gains of long-term learning from budget-matched restarts~\cite{wang2026sear}; PhysRSI-Eval~\cite{li2026physrsieval} measures such physical self-improvement of VLM agents on 18 RoboDojo tasks under a fixed interaction budget.
These systems output a reward, a policy, or code skills.
\method outputs a harness for a given VLM: it decides per stage whether a skill, a localizer, or a script is needed, trains the learned ones end to end (teacher RL, VLA distillation, keypoint networks), and revises them on the VLM's own failures.
ASPIRE's skills are programs and repair heuristics over the perception and control primitives it is given, whereas \method also trains the localizers that supply geometry no given interface provides.
ENPIRE improves a policy and its training code, and the resulting policy is deployed without the agent, whereas in \method the VLM stays in charge at deployment and the outer agent decides which tool takes each stage.
Concurrently, AgenticRoboRL~\cite{liu2026agenticroborl} post-trains the VLM planner itself with RL, and Teach and Grow~\cite{nie2026teachgrow} induces reusable skill blocks from demonstrations; both improve the planner but do not build perception tools.
Further related systems are discussed in \Cref{supp:related}.

\paragraph{VLAs and teacher--student distillation.}
Generalist VLAs are pretrained on large robot datasets and fine-tuned per task~\cite{black2024pi0,kim2025openvlaoft,nvidia2025gr00t,black2025pi05,zheng2025xvla}; hierarchical systems let a high-level model call low-level skills~\cite{ahn2022saycan}.
Privileged teachers trained in simulation are distilled into vision-based students with DAgger and noise injection~\cite{kim2025drs,laskey2017dart,ross2011dagger}.
In \method the VLA is one tool among several and is used only on stages it can learn; we measure where it stops working (it stops 4--9\,mm from the outlet on \texttt{plug\_in\_charger}, whose insertion requires about 2\,mm) and route those stages to geometry.

\paragraph{Geometric grounding for precise manipulation.}
Keypoint representations, pose estimation, and visual servoing are standard for insertion and assembly~\cite{chen2025trex,manuelli2019kpam,wen2024foundationpose}, and pointing VLMs and promptable segmentation give open-vocabulary localization.
On \texttt{plug\_in\_charger}, these generic tools are within 3\,mm on at most 51\% of the frames near the outlet, against 79\% for our keypoint localizer (\Cref{tab:localizer}).
Our localizers are trained by the outer agent from simulator labels on the states the deployed system reaches, and are exposed to the VLM as measurement tools rather than wired into a fixed controller.

\section{Method}
\label{sec:method}

\begin{figure*}[t]
  \centering
  \includegraphics[width=\linewidth]{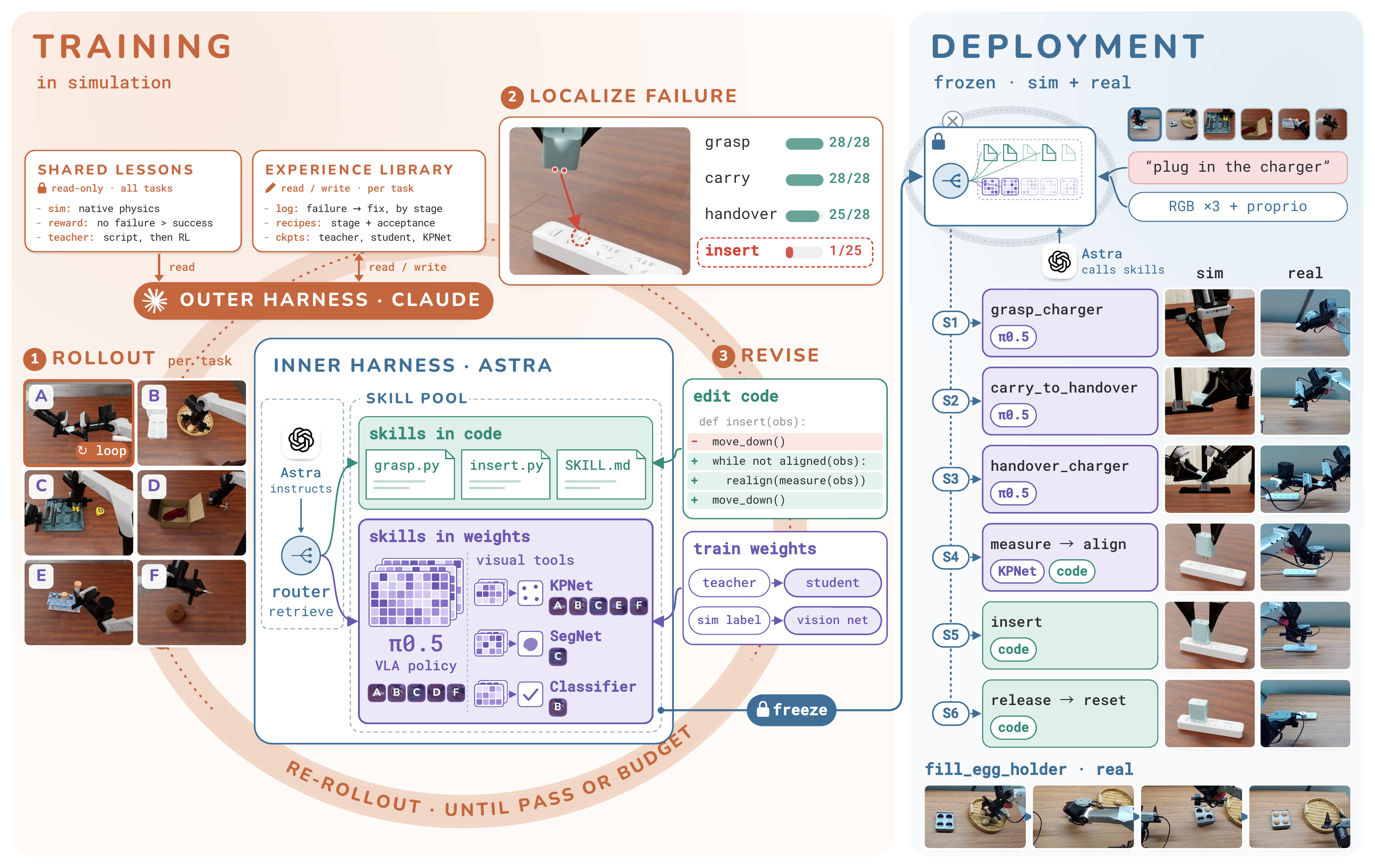}
  \caption{\textbf{Pipeline of \method.} In simulation, the outer harness (Claude) builds the inner harness of each task: it decomposes the task into stages, allocates a tool type to each stage with the takeover probe, and then rolls the inner harness out, localizes the first failing stage, and revises that stage, until every stage passes. It reads the shared lessons (the method note), which are the same for every task and never edited, and writes the experience library of the task. The frozen inner harness, Astra with a router over the built tools, is then deployed from RGB and proprioception alone, in simulation and on the real robot; on \texttt{plug\_in\_charger}, $\pi_{0.5}$ grasps, carries, and hands over the charger (S1--S3), and Astra aligns, inserts, and releases it with the KPNet offset tool (S4--S6).}
  \label{fig:overview}
\end{figure*}

\subsection{Problem Setting and Overview}
\label{sec:overview}

A task $\mathcal{T}$ is a language instruction together with a simulator, a layout distribution $p(\ell)$, and a success check; the robot observes three RGB cameras and proprioception.
The task is a chain of stages $S_1, \dots, S_n$ whose success probabilities multiply (\Cref{sec:outer}).

\paragraph{Two harnesses.}
\method is a harness that builds a harness (\Cref{fig:overview}).
The \emph{inner harness} $\mathcal{H}_{\text{in}} = (\mathcal{M}, \mathcal{R}, \mathcal{K})$ is what is deployed: a frozen VLM $\mathcal{M}$ (Astra), a router $\mathcal{R}$, and a set of tools $\mathcal{K}$ of three types, VLA skills, visual localizers, and scripts (\Cref{sec:inner}); the router tells the VLM which tools are available for the task, and at each stage the VLM calls a tool and checks the outcome in new images.
The baseline $(\mathcal{M}, \mathcal{K}_0)$ is the same VLM with only the base tools $\mathcal{K}_0$, a motion tool that executes a short chunk of end-effector commands per call.
The \emph{outer harness} $\mathcal{H}_{\text{out}} = (\mathcal{A}, \mathcal{E}, \mathcal{B})$ is an unmodified coding agent $\mathcal{A}$ (Claude) with an experience library $\mathcal{E}$ and build tools $\mathcal{B}$ (simulators, trainers, a GPU scheduler); the agent starts every task from the same method note, a generic training recipe that it reads but never edits, and writes $\mathcal{E}$ while it iterates.
It exists only at training time, in simulation, where it decomposes the task, allocates a tool type to each stage with the takeover probe, and then repeats three steps: \emph{rollout} of the inner harness on training layouts, \emph{localize failure} by estimating the success $\hat{p}_i$ of every stage and finding the first stage that breaks the chain, and \emph{revise} that stage, until all stages pass a gate $\gamma$ or the budget runs out (\Cref{sec:outer}).
The tools in weights are trained inside this loop (\Cref{sec:build}).

\paragraph{Stage limitations.}
A stage is \emph{perception-limited} if the VLM's estimate $\hat{x}_{\mathcal{M}}$ of the quantity $x$ it must control misses the tolerance $\epsilon$ by far, $\lVert \hat{x}_{\mathcal{M}} - x \rVert \gg \epsilon$, and \emph{control-limited} if the VLM chooses the correct action but cannot execute it reliably at its call rate; Astra, for instance, scores differently in three repeats of the same layout on 10 of 18 RoboDojo tasks.
The outer agent assigns a VLA skill to control-limited stages, a visual localizer to perception-limited stages, and the VLM itself, with a script where its behavior is routine, to the others.

\paragraph{RGB-only deployment.}
Privileged simulator state is used only for training, as teacher observations, labels, and for scoring.
The frozen inner harness reads only the three RGB cameras and proprioception, with fixed camera calibration and object CAD models as priors, and the same tools run in simulation and on the real robot (\Cref{sec:exp_main}).

\subsection{Inner Harness}
\label{sec:inner}

At deployment, Astra calls the tools in $\mathcal{K}$: it selects a tool, checks the outcome in the camera images, and decides the next step (\Cref{tab:tools}).

\paragraph{VLA skills.}
All VLA skills are served by one VLA student, $\pi_{0.5}$~\cite{black2025pi05}, conditioned on the stage instruction.
From the three camera images and proprioception, it predicts a chunk of $H$ joint-space actions, executes the first $h$, and predicts again ($H = 50$, $h = 5$; \Cref{supp:student}).
A skill runs one stage in closed loop without VLM calls and stops on a proprioceptive criterion or a step limit, \eg, grasp, carry, and handover on \texttt{plug\_in\_charger}.

\paragraph{Visual localizers.}
A visual localizer returns the metric quantity that a stage must control, together with a validity flag; it does not actuate the robot.
On \texttt{insert\_key}, \texttt{measure\_key\_offset} detects keypoints of the key and the lock in the wrist image, recovers both poses by PnP, and returns the tip-to-keyhole offset $\delta$; a reading is invalid when fewer than $N_{\min}$ keypoints are visible.
Keypoint networks are the default; segmentation networks measure regions, such as empty toolbox slots, and classifiers measure discrete states, such as whether an egg is seated.

\paragraph{Scripts.}
When the VLM applies the same rule to tool readings in every episode, the outer agent compiles the rule into a script that runs at the control rate.
For fine alignment, this rule is a guarded visual servo that corrects the measured offset $\delta_t$ at a fixed height and descends only within a funnel around the target:
\begin{align}
  g_t &= \mathbf{1}\big[\lVert \delta^{xy}_t \rVert < \epsilon_{xy} \,\wedge\, |\delta^{\text{yaw}}_t| < \epsilon_{\text{yaw}}\big], \label{eq:funnel}\\
  \Delta \mathbf{p}_t &= -K\,\delta^{xy,\text{yaw}}_t \;-\; g_t\, v_z\, \mathbf{e}_z , \label{eq:servo}
\end{align}
where $\Delta \mathbf{p}_t$ is the end-effector command, $K$ a gain, $v_z$ the descent speed, and $\mathbf{e}_z$ the vertical axis; after a failed insertion, the arm retracts and re-measures.
This rule runs as a script on \texttt{insert\_key} and \texttt{insert\_tubes}.
The VLM retains the decisions that no fixed rule covers: which stage comes next, whether a stage succeeded, and how to recover from unexpected states.
Tools return observations rather than judgments of success, so the VLM stays the judge of progress (\Cref{supp:interface}).
Besides the skill cards, which state each tool's function, precondition, cost, and measured reliability, the VLM reads task notes, which it writes itself by self-revision after the outer agent has frozen the tools (\Cref{supp:interface}).

\begin{table}[t]
  \centering
  \footnotesize
  \setlength{\tabcolsep}{2.5pt}
  \tabstyle
  \begin{adjustbox}{max width=\linewidth}\begin{tabular}{>{\raggedright\arraybackslash}p{0.15\linewidth}>{\raggedright\arraybackslash}p{0.25\linewidth}>{\raggedright\arraybackslash}p{0.32\linewidth}>{\raggedright\arraybackslash}p{0.22\linewidth}}
    \toprule
    \cellcolor{hsBlueBg}\hd{Task}
      & \cellcolor{hsPurpleBg}\grouphd{hsPurple}{VLA skills}
      & \cellcolor{hsPurpleBg}\grouphd{hsPurple}{Visual localizers}
      & \cellcolor{hsGreenBg}\grouphd{hsGreen}{Scripts} \\
    \midrule
    Plug & grasp, carry, handover & KPNet + PnP (wrist) & \\
    \altrow
    Key & grasp--lift, handover & templates (head); KPNet + PnP (wrist) & carry, insert--turn \\
    Tubes & & KPNet (wrist); SegNet & funnel insert \\
    \altrow
    Egg & & KPNet (head, wrist); classifier & grasp \\
    Toolbox & grasp, place & SegNet (head); KPNet (in-hand) & \\
    \altrow
    Pack & grasp, place & KPNet (wrist); SegNet & lid close \\
    Mugs & grasp, hang & KPNet (wrist) & \\
    \altrow
    Organize & grasp, place & SegNet (head) & \\
    \bottomrule
  \end{tabular}\end{adjustbox}
  \caption{\textbf{Inner harnesses built by the outer agent} in simulation. Per task, the stages run by VLA skills and by scripts, and the visual localizers used. KPNet: keypoint network; SegNet: segmentation network; parentheses: camera, or the held object for Toolbox. The VLM selects each tool and executes the stages not listed itself. Task abbreviations as in \Cref{tab:main_sim}.}
  \label{tab:tools}
\end{table}

\subsection{Outer Harness: Decompose, Allocate, Revise}
\label{sec:outer}

The outer agent builds the inner harness by three decisions, each based on stage success measured in rollouts: it decomposes the task into stages, allocates a tool type to each stage with the takeover probe, and revises the first stage that fails (\Cref{supp:alg:outer}).
The first two decisions set up the inner harness; the third is repeated until every stage passes the gate $\gamma$ or the budget is spent.

\paragraph{Decompose.}
The agent reads the task configuration and scoring code to recover the layout distribution and the tolerances, runs $(\mathcal{M}, \mathcal{K}_0)$ on a few layouts, and places stage boundaries where the failures in the transcripts cluster, with the official demonstrations as reference where available (\Cref{supp:fig:stages}).
A stage $S_i$ has a success predicate $c_i$ on measurable quantities, a failure predicate, and a step budget.
Since an episode succeeds only if every stage does,
\begin{equation}
  P(\text{success}) = \prod_{i=1}^{n} p_i, \qquad p_i = P(c_i \mid c_1, \dots, c_{i-1}),
  \label{eq:chain}
\end{equation}
the weakest stage bounds task success, and each stage can be built and evaluated on the state distribution induced by its predecessors.

\paragraph{Allocate.}
Each stage is assigned the tool type that addresses its limitation (\Cref{sec:overview}): a VLA skill if it is control-limited, a visual localizer read by the VLM or by a script if it is perception-limited, and the VLM with $\mathcal{K}_0$ otherwise.
For this, the agent trains a privileged teacher for every stage and distills a VLA student from them (\Cref{sec:build}).
The limitation is diagnosed by a \emph{takeover probe}, in which the student replaces the teacher on one stage of an otherwise successful chain.
If the chain still succeeds, the stage is learnable from images and is assigned to the VLA; if it fails although the teacher succeeds, the student cannot recover the controlled quantity from images at the required tolerance, and the stage is assigned a visual localizer.
On \texttt{plug\_in\_charger}, the alignment student stops several millimeters from the outlet and fails every insertion (0/28), whereas with the grasping, carrying, and handover students the chain still succeeds in 20/28 episodes (\Cref{sec:exp_diag}).
The agent then writes the router $\mathcal{R}$, which tells the VLM the tools of the task.

\paragraph{Revise.}
After each rollout of the inner harness in the native simulator, the agent estimates the success $\hat{p}_i$ of every stage and localizes the first stage below the gate $\gamma$, $i^\star = \min\{\, i : \hat{p}_i < \gamma \,\}$.
From the transcripts it attributes the failures of $S_{i^\star}$ to one of four causes, each of which determines the revision:
\emph{coverage}, a learned tool fails on entry states missing from its training data, so the agent resamples layouts and stage entries and retrains;
\emph{program}, a script fails on states it should handle, so the agent edits the code, \eg, by adding a regrasp when the wrist reaches its joint limit on \texttt{insert\_key};
\emph{interface}, the VLM misuses a working tool because of what the tool returns or what its card states, so the agent changes the interface, \eg, by removing success flags;
\emph{allocation}, the assigned tool type cannot overcome the stage's limitation, so the agent repeats the takeover probe and reassigns the stage.
Coverage, program, and interface failures are addressed by improving the current tool of $S_{i^\star}$, and allocation failures by training other tools for $S_{i^\star}$; the two run in parallel, and $S_{i^\star}$ receives the best of them, with $\mathcal{R}$ updated, only if $\hat{p}_{i^\star}$ rises in a paired comparison on identical layouts.

\paragraph{Experience library.}
The library $\mathcal{E}$ is what the agent writes while it iterates: it records the outcome of each decision, as text (stage recipes, acceptance protocols, failure--fix pairs) and as weights (teacher, student, and keypoint checkpoints), and it is empty at the start of a build from scratch.
The method note, in contrast, is fixed: the agent reads it but never edits it, and every build starts from the same note (\Cref{supp:outer}).
The note gives a generic training recipe, for the privileged teachers, their distillation into VLA students, and the localizers; every task-specific decision is the agent's: how to split the task into stages, which rewards and reference scripts to write, how to build the training replica, which tool type each stage receives, which data to record, and how to revise.

\subsection{Build: Training the Tools}
\label{sec:build}

Building is part of the allocation and of the revise loop rather than a separate phase.
Every learned tool is trained by the outer agent with the build tools $\mathcal{B}$: it builds the training scenes, writes the rewards and training configurations, launches and schedules the jobs, and accepts a checkpoint only if the stage passes the gate $\gamma$.
Training layouts are sampled uniformly from the full configured range of each task; the official layouts are not used to train or select any tool, weight, or script.

\paragraph{Privileged teachers.}
Every VLA skill starts from a teacher that observes the full simulator state, trained in a GPU-parallel state-based replica of the native scene from entry states that the previous stage actually reaches.
The agent writes a potential-based reward, and the teacher is trained with TD3~\cite{fujimoto2018td3} over action chunks, with its actor regularized toward an anchor policy that is first a rule-based reference controller and later the best policy found so far (\Cref{supp:teacher}, \Cref{supp:tab:td3}).

\paragraph{Student distillation.}
All accepted teachers are distilled into one $\pi_{0.5}$ student conditioned on the stage instruction, which sees only images and proprioception.
Data are collected in the native simulator with DAgger~\cite{ross2011dagger}, mixing teacher and student per action chunk and labeling every visited state with the teacher's clean action; the first iteration adds DART-style noise~\cite{laskey2017dart}, and only the action expert is trained (\Cref{supp:student}, \Cref{supp:tab:studenthp}).

\paragraph{Keypoint localizer.}
Visual localizers predict keypoints rather than masks.
A mask centroid shifts whenever the gripper occludes the object, which is the normal case near contact, whereas a few keypoints determine a 6-D pose even under partial occlusion.
Keypoints are defined on the object's CAD model, and their labels are the projections of these points under ground-truth poses, so no annotation is needed.
The network, KPNet, is a ResNet-18~\cite{he2015resnet} encoder--decoder that predicts one heatmap and one visibility logit per keypoint, and at test time the pose of a held object is
\begin{equation}
  \hat{T} = \arg\min_{T} \sum_{n:\, \hat{v}_n > 1/2} \big\lVert \hat{u}_n - \Pi(T X_n) \big\rVert^2,
  \label{eq:pnp}
\end{equation}
where $\hat{u}_n$ is the soft-argmax of heatmap $n$, $\hat{v}_n$ its visibility, $X_n$ the CAD keypoint, and $\Pi$ the calibrated projection.
KPNet has 11.5\,M parameters, trains in under 2\,h on one GPU, and reaches a median offset error of 0.13\,mm on frames with a true offset below 20\,mm, against 0.44\,mm for a head on frozen DINOv3~\cite{simeoni2025dinov3} features (\Cref{supp:localizer}, \Cref{supp:tab:dino}).

\section{Experiments}
\label{sec:exp}

\subsection{Experimental Setup}
\label{sec:exp_setup}

\paragraph{Benchmark and tasks.}
We use the RoboDojo simulation benchmark~\cite{chen2026robodojo}, bimanual ARX X5 tasks at 25\,Hz.
We evaluate on eight tasks, fixed before any experiment: the five tasks with the lowest best-entry score on the RoboDojo leaderboard of 2 November 2026, \texttt{fill\_egg\_holder}, \texttt{store\_tools\_in\_toolbox}, \texttt{pack\_objects\_into\_boxes}, \texttt{hang\_mugs}, and \texttt{organize\_table}, and three further insertion tasks that require fine-grained manipulation, \texttt{plug\_in\_charger}, \texttt{insert\_key}, and \texttt{insert\_tubes}.
They combine millimeter-level alignment (about 2\,mm and 3$^\circ$ before the plug is inserted), bimanual handovers, objects that slip or rotate in the grasp, and 300--900-step horizons.

\paragraph{Evaluation protocol.}
We follow the official protocol: 55 layouts for each of the official seeds 0, 1, 2, 165 episodes per task (on \texttt{fill\_egg\_holder}, 7 of the 195 layouts produce a broken initial scene and are discarded, as in the official evaluation, which leaves 188); \texttt{pack\_objects\_into\_boxes} and \texttt{hang\_mugs} have a second, random configuration with 30 layouts per seed (\texttt{hang\_mugs}: 30 base layouts per seed), and on these two tasks we report, as the leaderboard does, the mean of the base and random layouts (\Cref{supp:eval}).
Each task is evaluated with the final inner harness, frozen before the evaluation: its tools and router are built by the outer agent, and its task notes are written by the VLM through self-revision on self-generated layouts (\Cref{supp:interface}).
The outer agent develops and accepts every change on layouts that it samples from the task configuration; official layouts are used only to test frozen versions, here and in the builds from scratch (\Cref{sec:exp_build}); each frozen version is tested once, and the result is not returned to the agent.
Consistently, success on official seed~1, on which every version of the builds from scratch is tested, does not differ significantly from that on seeds 0 and 2 (53/55 vs.\ 108/110 on \texttt{plug\_in\_charger}, 51/55 vs.\ 106/110 on \texttt{insert\_key}, 34/63 vs.\ 58/125 on \texttt{fill\_egg\_holder}; Fisher's exact test, $p\geq0.36$).
The metric is native success on the first attempt, one episode per layout from the initial scene; timeouts, refusals, and episodes that fail at reset count as failures.

\paragraph{Baselines and implementation.}
We compare with the best entry per task on the public RoboDojo leaderboard~\cite{chen2026robodojo,community2026xpolicylab} as of 2 November 2026 (VLA/WAM or agentic); Astra, direct control $(\mathcal{M}, \mathcal{K}_0)$, the organizers' reference harness, \ie, the same VLM with only the base tools, so the difference from \method is attributable to the built tools; the tool-based harnesses Auriga~\cite{wang2026auriga} and PyRUA-Lean~\cite{si2026pyrualean}; and the VLA $\pi_{0.5}$~\cite{black2025pi05}.
The base VLM is GPT-6 Astra (\texttt{xhigh}) and the outer agent is Claude Opus 5.5 (\texttt{high}); further details are in \Cref{supp:eval,supp:build}.

\paragraph{Questions.}
The experiments answer four questions, which test the three contributions: does the built harness solve fine-grained tasks, in simulation and on the real robot (\Cref{sec:exp_main})? Does each stage need the tool type that the takeover probe assigns (\Cref{sec:exp_diag})? Can the outer agent build the harness from scratch (\Cref{sec:exp_build})? What does the built harness cost at deployment (\Cref{sec:exp_eff})?

\begin{table*}[t]
  \centering
  \footnotesize
  \setlength{\tabcolsep}{4pt}
  \tabstyle
  \begin{adjustbox}{max width=\linewidth}\begin{tabular}{l ccccc>{\columncolor{hsBlueBg!55}}c ccc>{\columncolor{hsBlueBg!55}}c >{\columncolor{hsBlueBg}}c}
    \toprule
    \headrow
    & \multicolumn{6}{c}{\hd{Hardest five}} & \multicolumn{4}{c}{\hd{Insertion}} & \\
    \cmidrule(lr){2-7} \cmidrule(lr){8-11}
    \headrow
    \hd{Method} & \hd{Egg} & \hd{Toolbox} & \hd{Pack} & \hd{Mugs} & \hd{Organize} & \hd{Avg.} & \hd{Plug} & \hd{Key} & \hd{Tubes} & \hd{Avg.} & \hd{All} \\
    \midrule
    Leaderboard best (2 Nov 2026)        & 6.0 & 10.0 & 30.0 & 32.0 & 36.0 & 22.8 & 70.0 & 32.0 & 92.0 & 64.7 & 38.5 \\
    Astra, direct control~\cite{chen2026robodojo} & 0.0 & 0.0 & 0.0 & 0.0 & 0.0 & 0.0 & 0.0 & 0.0 & 0.0 & 0.0 & 0.0 \\
    Auriga~\cite{wang2026auriga}         & 0.7 & 0.0 & 14.7 & 4.0 & 19.3 & 7.7 & 20.7 & 0.0 & 46.0 & 22.2 & 13.2 \\
    PyRUA-Lean~\cite{si2026pyrualean}    & 6.0 & 0.0 & 30.0 & 12.0 & 4.0 & 10.4 & 16.0 & 0.0 & 50.0 & 22.0 & 14.8 \\
    $\pi_{0.5}$~\cite{black2025pi05}     & 0.0 & 0.0 & 2.7 & 0.0 & 0.0 & 0.5 & 0.0 & 0.0 & 3.3 & 1.1 & 0.8 \\
    \midrule
    \oursrow
    \method (ours)                       & \textbf{48.9} & \textbf{79.4} & \textbf{47.4} & \textbf{41.1} & \textbf{47.3} & \textbf{52.8} & \textbf{97.6} & \textbf{95.2} & \textbf{95.2} & \textbf{96.0} & \textbf{69.0} \\
    \bottomrule
  \end{tabular}\end{adjustbox}
  \caption{\textbf{First-attempt success rate (\%, $\uparrow$)} on all official layouts of each task: 165 (3 seeds $\times$ 55), 188 for \texttt{fill\_egg\_holder}; for Pack and Mugs, the mean of base and random layouts, as on the leaderboard (Pack: 165 base and 90 random; Mugs: 90 and 90). Plug: \texttt{plug\_in\_charger}; Key: \texttt{insert\_key}; Tubes: \texttt{insert\_tubes}; Egg: \texttt{fill\_egg\_holder}; Toolbox: \texttt{store\_tools\_in\_toolbox}; Pack: \texttt{pack\_objects\_into\_boxes}; Mugs: \texttt{hang\_mugs}; Organize: \texttt{organize\_table}. Hardest five: Egg, Toolbox, Pack, Mugs, Organize; Insertion: Plug, Key, Tubes. Avg.\ is the mean over the tasks of the group, All over all eight; best per column in bold. Baselines are from the official RoboDojo leaderboard of 2 Nov 2026, with 150 episodes per task (Auriga: from its report); per-seed results in \Cref{supp:tab:seeds}, ratios to the best entry in \Cref{supp:tab:gains}, protocol in \Cref{supp:eval}.}
  \label{tab:main_sim}
\end{table*}

\subsection{Does the Built Harness Solve Fine-Grained Tasks?}
\label{sec:exp_main}

\paragraph{Simulation.}
\Cref{tab:main_sim} gives the main results.
Over all official episodes, \method reaches 97.6\% on \texttt{plug\_in\_charger}, 95.2\% on both \texttt{insert\_key} and \texttt{insert\_tubes}, and 69.0\% on average over the eight tasks, against 38.5\% for the best leaderboard entry per task as of 2 Nov 2026, 13.2\% for Auriga, and 14.8\% for PyRUA-Lean; on the five hardest tasks the averages are 52.8\% vs.\ 22.8\% for the best entries, and \method is best on every task.
\method also outperforms the methods trained for each task: the VLA entries of the leaderboard, including $\pi_{0.5}$ (0.8\% on average), are fine-tuned on simulated demonstrations of each task.
Astra under direct control succeeds on none of the eight tasks: at 5 control steps per VLM call, repeated attempts at the stages that require precise alignment or stable contact use up the step limit of the episode.
Since the same VLM is the base model of \method, the gap is attributable to the built tools.
Success is lower on the tasks that involve many objects or free-hanging contacts, \texttt{hang\_mugs} (41.1\%), \texttt{fill\_egg\_holder} (48.9\%), \texttt{organize\_table} (47.3\%), and \texttt{pack\_objects\_into\_boxes} (47.4\%); on base and random layouts separately, \method reaches 58.2\% and 36.7\% on \texttt{pack\_objects\_into\_boxes} and 54.4\% and 27.8\% on \texttt{hang\_mugs} (\Cref{supp:tab:seeds}).
The largest ratio to the best entry is 8.2$\times$, on \texttt{fill\_egg\_holder} (\Cref{supp:tab:gains}).

\paragraph{Real robot.}
The real setup is an ARX X5 pair with a head camera and two wrist RGB cameras, calibrated intrinsics, mounts, and joint zero offsets (median held-out reprojection error of 0.84\,px on the right wrist).
The task objects are 3D prints of the simulation assets, so the CAD geometry used by the localizers is unchanged.
The deployed tools are the same as in simulation; skills run with real-time chunking~\cite{black2025rtc}.
On 10 trials per task (\Cref{tab:real}), the deployed harness succeeds in 9/10 trials on \texttt{plug\_in\_charger} and \texttt{insert\_key} and in 5/10 on \texttt{fill\_egg\_holder}.
Deployment is zero-shot: the VLA and the keypoint network are trained only in simulation and kept frozen, and no real-robot demonstration is used.
As a reference trained on real data, $\pi_{0.5}$ post-trained on 100 teleoperated demonstrations per task succeeds in 2/10, 0/10, and 5/10 trials: as often as the harness on \texttt{fill\_egg\_holder}, but rarely or never on the two tasks that require millimeter-level alignment after a handover (\Cref{supp:real}).

\begin{table}[t]
  \centering
  \footnotesize
  \setlength{\tabcolsep}{4pt}
  \tabstyle
  \begin{adjustbox}{max width=\linewidth}\begin{tabular}{lccc}
    \toprule
    \headrow
    & \hd{Plug} & \hd{Key} & \hd{Egg} \\
    \midrule
    \multicolumn{4}{>{\columncolor{hsBlueBg!55}}l}{\grouphd{hsBlue}{\emph{Success} $\uparrow$}} \\
    \oursrow
    \method (0 real demos)                        & \textbf{9/10} & \textbf{9/10} & \textbf{5/10} \\
    $\pi_{0.5}$ (100 teleop.\ demos per task)    & 2/10 & 0/10 & \textbf{5/10} \\
    \midrule
    \multicolumn{4}{>{\columncolor{hsBlueBg!55}}l}{\grouphd{hsBlue}{\emph{Time per trial} (s) $\downarrow$}} \\
    \oursrow
    \method                                       & \textbf{100} & \textbf{137} & \textbf{219} \\
    Direct VLM control (estimated)                & 1475 & 1110 & 3225 \\
    \rowcolor{hsGreenBg}
    \grouphd{hsGreen}{Speed-up}                   & \grouphd{hsGreen}{14.8$\times$} & \grouphd{hsGreen}{8.1$\times$} & \grouphd{hsGreen}{14.7$\times$} \\
    \bottomrule
  \end{tabular}\end{adjustbox}
  \caption{\textbf{Real-robot results}, 10 trials per task. \method is deployed zero-shot from simulation; $\pi_{0.5}$ is post-trained on teleoperated demonstrations on the same setup. 95\% Wilson intervals: 59.6--98.2 (9/10), 23.7--76.3 (5/10), 5.7--51.0 (2/10), 0--27.8 (0/10); \method vs.\ $\pi_{0.5}$, Fisher exact test: $p{=}0.005$ (Plug), $p{<}10^{-3}$ (Key). Direct VLM control: estimated from the control steps of our successful simulated episodes (295, 222, 645) at one VLM call per 5 steps and 25\,s per call (\Cref{supp:efficiency}). Plug: \texttt{plug\_in\_charger}; Key: \texttt{insert\_key}; Egg: \texttt{fill\_egg\_holder}.}
  \label{tab:real}
\end{table}

\begin{table}[t]
  \centering
  \footnotesize
  \tabstyle
  \begin{adjustbox}{max width=\linewidth}\begin{tabular}{llc}
    \toprule
    \headrow
    \hd{S1--S3 by} & \hd{S4--S5 by} & \hd{Full chain} \\
    \midrule
    Teacher & Teacher & 28/28 \\
    \altrow
    Student & Teacher & 20/28 \\
    Teacher & Student & 0/28 \\
    \bottomrule
  \end{tabular}\end{adjustbox}
  \caption{\textbf{Takeover probe} on \texttt{plug\_in\_charger} (S1--S3: grasp, carry, handover; S4--S5: align, insert), 14 handover layouts $\times$ 2. Rows 2--3 use an earlier student (25/28 handover).}
  \label{tab:takeover}
\end{table}

\subsection{Does Each Stage Need the Tool It Is Given?}
\label{sec:exp_diag}

\paragraph{The takeover probe separates the two limitations.}
\Cref{tab:takeover} reports the takeover probe on \texttt{plug\_in\_charger}.
With the student on grasp, carry, and handover, the chain still succeeds in 20/28 episodes, whereas with the student on alignment and insertion it succeeds in none: the student stops 4--9\,mm from the outlet, descends, and hits the strip.
Unfreezing the vision encoder does not help: trained from the same student on identical data, the frozen and unfrozen variants reach 2/28 and 3/28 (\Cref{supp:student}).
On the insertion stage of \texttt{insert\_key}, the same probe gives 3/32 for the student against 26/28 for the teacher.
The first three stages are therefore control-limited and the last two perception-limited, and the outer agent assigns them a VLA skill and a keypoint localizer.

\paragraph{Each tool type is needed where it is assigned.}
\Cref{tab:ablation} removes one type of tool at a time from the final inner harness; the VLM then executes the affected stages with the remaining tools.
Without visual localizers, the VLM must align from the RGB images, and success drops to 20.0\%, 3.6\%, and 0.0\% on \texttt{plug\_in\_charger}, \texttt{insert\_key}, and \texttt{insert\_tubes}.
Without VLA skills, the contact-rich stages return to the VLM at 5 control steps per call, and success drops to 9.1\% and 0.0\%.
Without scripts, success drops to 0.0\% on \texttt{insert\_key}, whose scripts correct the key at every control step, but stays at 85.5\% on \texttt{insert\_tubes}, whose tolerance is coarse (exact McNemar test against the full harness, $p{=}0.29$).
Removing the reliability statistics from the skill cards changes nothing (53/55 on the same layouts).
Failure modes are analyzed in \Cref{supp:toolabl}.
Assigning a stage to another tool type than the probe chose thus loses most of the success, except where the tolerance is coarse enough for the VLM.

\begin{figure*}[t]
  \centering
  \includegraphics[width=\linewidth]{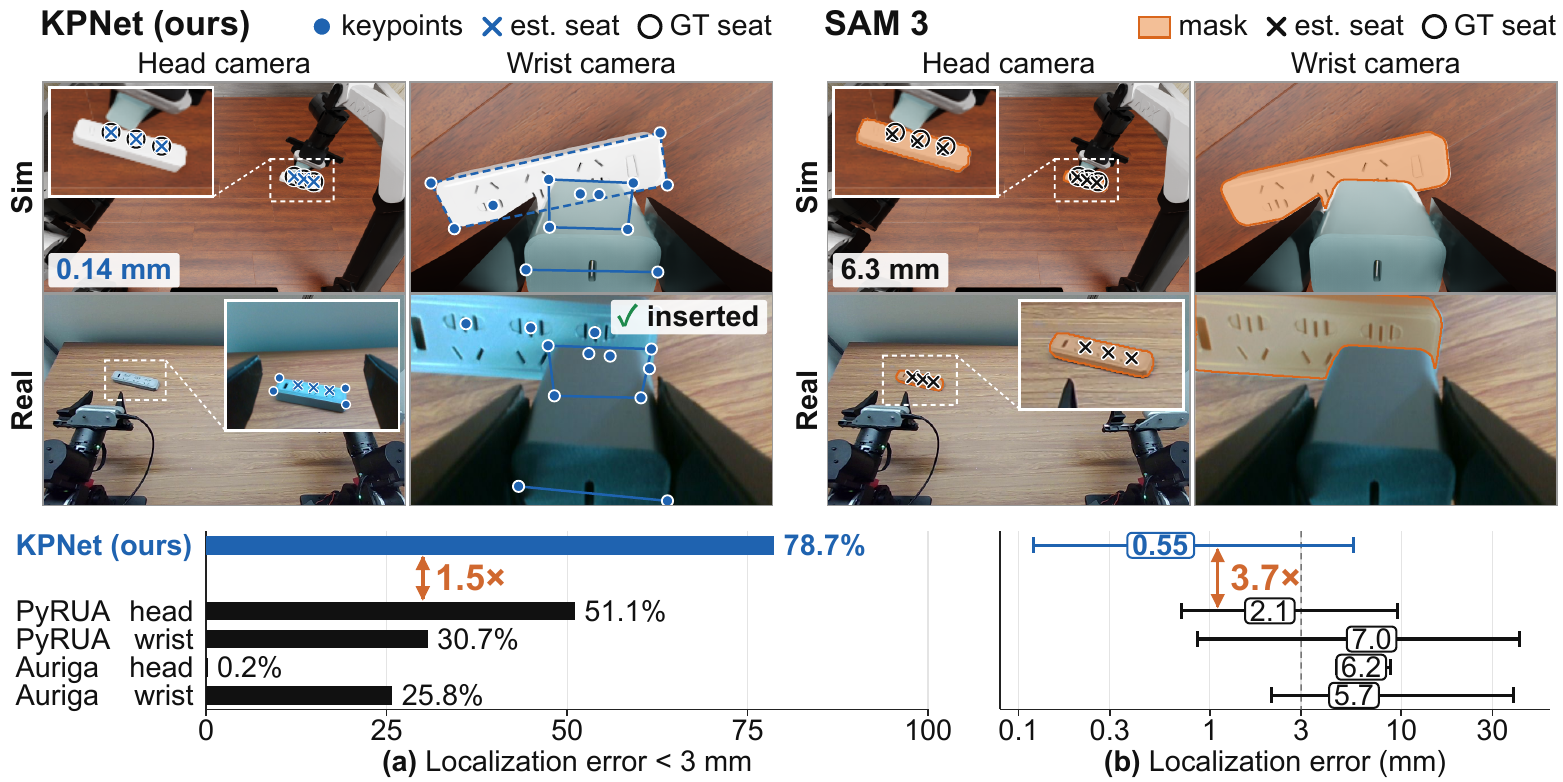}
  \caption{\textbf{Keypoint localizer vs.\ SAM3-based localization} on \texttt{plug\_in\_charger}. Top: frames from the head and wrist cameras in simulation and on the real robot. KPNet predicts keypoints on the power strip (dots) and recovers the outlet positions ($\times$) by PnP; SAM3 segments the strip (orange), and the outlet positions follow from the mask. Circles mark the ground-truth outlets; numbers give the mean outlet error on the simulated frame. Bottom: (a)~fraction of near frames (gripper tip within 6\,cm of the strip) with an error below 3\,mm, where a frame without an estimate counts as a failure; (b)~median error and 10th--90th percentile range over the frames with an estimate (log scale). PyRUA-Lean and Auriga localize from SAM3 masks; Auriga gets ground-truth depth. Frames and metric as in \Cref{tab:localizer}.}
  \label{fig:loc_compare}
\end{figure*}

\paragraph{A trained keypoint localizer, not generic perception, supplies the missing geometry.}
\Cref{tab:localizer} compares the keypoint localizer with the SAM3-based~\cite{carion2026sam3} perception of the two tool-based baselines on \texttt{plug\_in\_charger}, on every localizer call the VLM made after the handover; all methods receive the same RGB images, and Auriga, in its favor, gets ground-truth depth; \Cref{fig:loc_compare} shows example frames (protocol in \Cref{supp:loccmp}).
Over all these calls, the keypoint localizer has a median error of 0.5\,mm, and it is within 3\,mm on 78.7\% of the near frames, the accuracy target we set for the offset measurement, against 51.1\% for PyRUA-Lean and at most 25.8\% for Auriga.
Even with true depth, Auriga is biased by about 6\,mm from the head camera, because the mask includes the visible side faces of the strip and the fitted center shifts toward the camera, so the error lies in the mask-to-geometry step, not in depth; even on the ground-truth silhouette of the strip, the best mask any trained segmenter could produce, Auriga's geometry errs by 4.9\,mm from the head camera and 2.1\,mm from the wrist camera (median), against 0.5\,mm for KPNet (\Cref{supp:tab:newframes}).
Both sides of this comparison are trained, so it isolates the choice of keypoints over masks rather than training itself.
From the wrist camera both baselines see only part of the strip and their p90 error approaches 40\,mm, whereas keypoints with PnP on the known CAD model handle the partial view; when the strip is not in view, the localizer reports a failure, not a wrong reading.
On 1212 new frames from a rerun on the same 110 layouts, KPNet is within 3\,mm on 65.1\% of all frames, against at most 45.2\% for PyRUA-Lean and 56.7\% for Auriga's geometry on the ground-truth mask of the strip (\Cref{supp:tab:newframes}).
All these frames come from trajectories of the KPNet-based harness; when KPNet is replaced by Auriga's localizer in the deployed harness, success on seed~0 drops from 52/55 to 14/55 (exact McNemar test, $p{<}10^{-10}$).

\begin{table}[t]
  \centering
  \footnotesize
  \setlength{\tabcolsep}{4pt}
  \tabstyle
  \begin{adjustbox}{max width=\linewidth}\begin{tabular}{lccc}
    \toprule
    \headrow
    \hd{Variant} & \hd{Plug} & \hd{Key} & \hd{Tubes} \\
    \midrule
    \oursrow
    \method (full inner harness)                  & \textbf{96.4} & \textbf{95.2} & \textbf{95.2} \\
    \midrule
    \multicolumn{4}{>{\columncolor{hsBlueBg!55}}l}{\grouphd{hsBlue}{\emph{Tools}}} \\
    \quad w/o VLA skills                          & 9.1  & 0.0 & -- \\
    \altrow
    \quad w/o visual localizers                   & 20.0 & 3.6 & 0.0 \\
    \quad w/o scripts                             & --   & 0.0 & 85.5 \\
    \midrule
    \multicolumn{4}{>{\columncolor{hsBlueBg!55}}l}{\grouphd{hsBlue}{\emph{Interface}}} \\
    \quad skill cards w/o measured reliability    & 96.4 & --  & -- \\
    \bottomrule
  \end{tabular}\end{adjustbox}
  \caption{\textbf{Ablations} on official layouts (first-attempt success, \%). Every variant removes one part of the final inner harness and runs on all 55 layouts of seed~0 with the same initial states and step budgets. Full inner harness: Plug re-run on all 165 layouts (159/165, 53/55 on seed~0; \Cref{tab:main_sim} reports 161/165 from the main run), Key and Tubes as in \Cref{tab:main_sim}. On Plug, removing the VLA skills or the localizers hurts significantly against the full harness on the same layouts (exact McNemar test, $p{<}10^{-9}$). --: the harness has no such tool (\Cref{tab:tools}).}
  \label{tab:ablation}
\end{table}

\begin{table}[t]
  \centering
  \footnotesize
  \setlength{\tabcolsep}{3pt}
  \tabstyle
  \begin{adjustbox}{max width=\linewidth}\begin{tabular}{lcccccc}
    \toprule
    \headrow
    & & \multicolumn{3}{c}{\hd{All frames}} & \hd{Near} & \\
    \cmidrule(lr){3-5} \cmidrule(lr){6-6}
    \headrow
    \hd{Method} & \hd{Found} & \hd{Med.\ } & \hd{p90} & \hd{${<}3$\,mm} & \hd{${<}3$\,mm} & \hd{Time} \\
    \headrow
    & \hd{(\%)} & \hd{(mm)} & \hd{(mm)} & \hd{(\%) $\uparrow$} & \hd{(\%) $\uparrow$} & \hd{(s)} \\
    \midrule
    PyRUA-Lean (head)       & 74.3 & 2.1 & 9.6  & 47.6 & 51.1 & 3.3 \\
    \altrow
    PyRUA-Lean (wrist)       & 84.7 & 7.0 & 41.4 & 25.5 & 30.7 & 2.0 \\
    Auriga$^\dagger$ (head)       & 71.0 & 6.2 & 8.8  & 0.2  & 0.2  & 0.3 \\
    \altrow
    Auriga$^\dagger$ (wrist)       & 78.4 & 5.7 & 38.8 & 21.3 & 25.8 & 0.4 \\
    \midrule
    \oursrow
    KPNet + PnP (ours)       & 78.7 & \textbf{0.5} & \textbf{5.6} & \textbf{67.2} & \textbf{78.7} & \textbf{0.13} \\
    \bottomrule
  \end{tabular}\end{adjustbox}
  \caption{\textbf{Localizer accuracy} on \texttt{plug\_in\_charger}: mean $xy$ error of the three outlet positions over 1174 frames at which the VLM called the localizer after the handover (official seeds 0 and 2); Near: 963 frames with the gripper tip within 6\,cm of the power strip. Head/wrist: camera used; ours uses the wrist camera. Median and p90 are over frames with an estimate; the ${<}3$\,mm rates count missing estimates as failures. $^\dagger$With ground-truth depth instead of VGGT and SAM3 instead of SAM3.1, which gives nearly the same results (\Cref{supp:tab:newframes}); for both baselines, the $180^\circ$ ambiguity of the strip is resolved toward the ground truth.}
  \label{tab:localizer}
\end{table}

\begin{table}[t]
  \centering
  \footnotesize
  \setlength{\tabcolsep}{2.5pt}
  \tabstyle
  \begin{adjustbox}{max width=\linewidth}\begin{tabular}{lcccccc}
    \toprule
    \headrow
    & & \multicolumn{3}{c}{\hd{All frames}} & \hd{Near} & \\
    \cmidrule(lr){3-5} \cmidrule(lr){6-6}
    \headrow
    \hd{Method} & \hd{Found} & \hd{Med.} & \hd{p90} & \hd{${<}3$\,mm} & \hd{${<}3$\,mm} & \hd{Time} \\
    \headrow
    & \hd{(\%)} & \hd{(mm)} & \hd{(mm)} & \hd{(\%)} & \hd{(\%)} & \hd{(s)} \\
    \midrule
    PyRUA-Lean (h) & 70.5 & 2.0 & 10.7 & 45.2 & 48.6 & 2.7 \\
    \altrow
    PyRUA-Lean (w) & 83.7 & 5.7 & 39.4 & 27.7 & 32.9 & 1.7 \\
    Auriga$^\dagger$ SAM3 (h) & 69.1 & 6.2 & 9.7 & 0.4 & 0.5 & 0.3 \\
    \altrow
    Auriga$^\dagger$ SAM3 (w) & 77.5 & 5.4 & 34.5 & 20.7 & 25.1 & 0.4 \\
    Auriga$^\dagger$ SAM3.1 (h) & 69.5 & 6.2 & 9.6 & 0.4 & 0.5 & 0.2 \\
    \altrow
    Auriga$^\dagger$ SAM3.1 (w) & 77.6 & 5.3 & 33.4 & 20.5 & 24.9 & 0.3 \\
    Auriga$^\dagger$ true mask (h) & 100.0 & 4.9 & 6.4 & 0.0 & 0.0 & -- \\
    \altrow
    Auriga$^\dagger$ true mask (w) & 92.0 & 2.1 & 16.4 & 56.7 & 66.4 & -- \\
    \midrule
    \oursrow
    KPNet + PnP (ours) & 77.8 & \textbf{0.5} & \textbf{4.6} & \textbf{65.1} & \textbf{76.5} & \textbf{0.1} \\
    \bottomrule
  \end{tabular}\end{adjustbox}
  \caption{\textbf{Localizer accuracy on new frames} of \texttt{plug\_in\_charger}: all 1212 localizer calls after the handover in a rerun of the deployed harness on the 110 official layouts of seeds 0 and 2; Near: 996 frames with the gripper tip within 6\,cm of the power strip. Columns as in \Cref{tab:localizer}; h/w: head/wrist camera. $^\dagger$Ground-truth depth instead of VGGT; true mask: the ground-truth silhouette of the power strip instead of the segmentation.}
  \label{supp:tab:newframes}
\end{table}

\subsection{Can the Outer Agent Build the Harness?}
\label{sec:exp_build}

\paragraph{Building from scratch.}
Starting from the method note with an empty experience library, without code or checkpoints of earlier builds, the outer agent builds the inner harness of \texttt{plug\_in\_charger} to 94.5\% on official seed~1 after 7.1\,h with one human intervention, and those of \texttt{insert\_key} (90.9\% after 4.5\,h), \texttt{hang\_mugs}, and \texttt{fill\_egg\_holder} without intervention (\Cref{supp:tab:builds}).
Each build exceeds the best leaderboard entry within hours, and the revise loop raises it further, \eg, \texttt{insert\_key} from 50.9\% for its first tested version to 94.5\% (\Cref{supp:tab:keybuild}).
With longer iteration, the builds from scratch come close to the evaluated harnesses: on the same official seed~1, they reach 96.4\% on \texttt{plug\_in\_charger} after 12.2\,h, 94.5\% on \texttt{insert\_key} after 8.0\,h, and 47.6\% on \texttt{fill\_egg\_holder} after 12.1\,h, against 96.4\% (53/55), 92.7\% (51/55), and 54.0\% (34/63) for the evaluated harnesses.
Every decision of the outer agent leaves a tool in the inner harness, so the simpler alternatives to its decisions are the inner-harness variants of \Cref{sec:exp_diag} and \Cref{supp:tab:outeralt} (\Cref{supp:outer}).

\paragraph{The revise loop.}
On \texttt{plug\_in\_charger}, a coverage revision raises the handover student from 22/25 with behavior cloning to 28/28 (197/200 on unseen layouts) after DAgger and full-range layouts (\Cref{supp:student}).
On \texttt{insert\_key}, three revise rounds raise official success from 145 through 148 and 153 to 157 of 165; each change is accepted on a paired sign test over training layouts before the official layouts are evaluated.
On \texttt{pack\_objects\_into\_boxes}, whose random layouts use object models that never appear in the base layouts or in the training of its localizer, the outer agent added two modules for the random-layout failures, localizer training on closed-loop mid-episode frames and box re-alignment; these modules also raise success on the base layouts, from 83/165 to 96/165 (\Cref{supp:eval}).

\begin{table}[t]
  \centering
  \footnotesize
  \setlength{\tabcolsep}{3pt}
  \tabstyle
  \begin{adjustbox}{max width=\linewidth}\begin{tabular}{lccc}
    \toprule
    \headrow
    \hd{Task} & \hd{Time} & \hd{SR} & \hd{Human} \\
    \midrule
    Plug & 7.1\,h & 94.5 & 1 \\
    \altrow
    Key & 4.5\,h & 90.9 & 0 \\
    Mugs & 12.8\,h & 52.0 / 32.0 & 0 \\
    \altrow
    Egg & 3.0\,h & 46.0 & 0 \\
    \bottomrule
  \end{tabular}\end{adjustbox}
  \caption{\textbf{Builds from scratch.} Time: wall clock until the reported version, excluding the periods in which the outer agent had stopped and waited to be resumed; SR: its success rate (\%); Human: human interventions until then. Each build exceeds the best leaderboard entry within hours, while the highest success needs longer iteration, \eg, 12.2\,h on Plug (96.4\% on official seed~1), 8.0\,h on Key (94.5\%), and 12.1\,h on Egg (47.6\%). Mugs SR is given for the base and random layouts separately, on 25 layouts.}
  \label{supp:tab:builds}
\end{table}

\begin{table}[t]
  \centering
  \footnotesize
  \setlength{\tabcolsep}{4pt}
  \tabstyle
  \begin{adjustbox}{max width=\linewidth}\begin{tabular}{lccc}
    \toprule
    \headrow
    \hd{Method} & \hd{VLM calls $\downarrow$} & \hd{Time (s) $\downarrow$} & \hd{Succ.\ (\%) $\uparrow$} \\
    \midrule
    Astra, direct control         & 80 & 1956 & 0.0 \\
    \oursrow
    \method (ours)                & 31 & 1049 & 96.4 \\
    \midrule
    \rowcolor{hsGreenBg}
    \grouphd{hsGreen}{Reduction vs.\ direct} & \grouphd{hsGreen}{2.6$\times$} & \grouphd{hsGreen}{1.9$\times$} & -- \\
    \bottomrule
  \end{tabular}\end{adjustbox}
  \caption{\textbf{Deployment efficiency} on \texttt{plug\_in\_charger} with the same VLM (Astra). Medians per episode of the number of VLM calls and of the time to success, where a failed episode counts as the time at the episode limit. Direct control never succeeds, so its time is the time to reach that limit (\Cref{supp:efficiency}).}
  \label{tab:efficiency}
\end{table}

\subsection{What Does the Harness Cost at Deployment?}
\label{sec:exp_eff}

Under direct control, every 5 control steps cost one VLM call of about 25\,s, so a \texttt{plug\_in\_charger} episode reaches the 400-step limit after 80 calls and about 33\,min without succeeding.
In \method, each skill call executes a whole stage in closed loop at the control rate, and scripts run whole stages without the VLM.
\Cref{tab:efficiency} reports VLM calls per episode and time to success, capped at the episode limit since the baseline never succeeds.
On \texttt{plug\_in\_charger}, \method needs 2.6$\times$ fewer VLM calls and 1.9$\times$ less time; the grasp, alignment, and insertion take only 252\,s, and most of the remaining time is the final arm reset, which the VLM still drives chunk by chunk.
On the real robot, an uncut trial takes 100\,s, 137\,s, and 219\,s on \texttt{plug\_in\_charger}, \texttt{insert\_key}, and \texttt{fill\_egg\_holder} (\Cref{tab:real}).
Direct VLM control is not run on the real robot; estimated from the control steps of our successful simulated episodes, it would take 8--15$\times$ longer per trial, a lower bound since it does not complete these tasks in simulation (\Cref{supp:efficiency}).

\section{Conclusion}
\label{sec:conclusion}

VLM harnesses fail at fine-grained manipulation on the stages whose geometry the VLM cannot perceive or whose control it cannot execute, and a single such stage bounds the whole task.
Which of the two a stage lacks can be measured with a takeover probe, and the measurement determines the tool to build: a VLA skill, a visual localizer, or a script.
\method hands these measurements and the building to an unmodified coding agent with a fixed method note, which revises the first failing stage until every stage passes.
The resulting harness reaches 95.2--97.6\% on the three insertion tasks and 69.0\% on average over eight fine-grained RoboDojo tasks under the full official protocol, transfers zero-shot to the real robot (23 of 30 trials), and needs 2.6$\times$ fewer VLM calls than direct VLM control on \texttt{plug\_in\_charger}.
The broader point is that the harness, not only the model, can be built automatically, against the failures of the model that will use it.

\paragraph{Limitations.}
\method needs a task simulator and object CAD models to train the object-specific localizers, which limits deployment to tasks with known geometry. Asynchronous reasoning remains an open challenge that future work could address. Finally, the method has yet to be validated across different embodiments.

{
  \small
  \bibliographystyle{ieeenat_fullname}
  \bibliography{main_arxiv}
}

\clearpage
\beginappendix
\suppressfloats[t]% no float above the appendix heading
% Appendix of the arXiv version: supplementary material of the CVPR submission #6978,
% content aligned with sec8/X_supp_clean.tex (2026-11-29); the main text cites each appendix item.
\section{Evaluation Details}
\label{supp:eval}

\paragraph{Baselines.}
Baseline numbers are taken from the public RoboDojo leaderboard~\cite{chen2026robodojo,community2026xpolicylab} as of 2 November 2026, except for Auriga~\cite{wang2026auriga}, whose numbers are taken from its report.
The best leaderboard entry per task is PyRUA-Lean~\cite{si2026pyrualean} on \texttt{fill\_egg\_holder} and \texttt{pack\_objects\_into\_boxes}, GO-3 Nano on \texttt{store\_tools\_in\_toolbox}, \texttt{plug\_in\_charger}, \texttt{insert\_key}, and \texttt{insert\_tubes}, VPP2 on \texttt{hang\_mugs}, and Q0-Preview on \texttt{organize\_table}.
\Cref{supp:tab:gains} lists, per task, the ratio to the best entry.

\begin{table}[t]
  \centering
  \footnotesize
  \setlength{\tabcolsep}{4pt}
  \tabstyle
  \begin{adjustbox}{max width=\linewidth}\begin{tabular}{lccc}
    \toprule
    \headrow
    \hd{Task} & \hd{Ours} & \hd{Best} & \hd{Ratio} \\
    \midrule
    Egg      & 48.9 & 6.0  & 8.2$\times$ \\
    \altrow
    Toolbox  & 79.4 & 10.0 & 7.9$\times$ \\
    Pack     & 47.4 & 30.0 & 1.6$\times$ \\
    \altrow
    Mugs     & 41.1 & 32.0 & 1.3$\times$ \\
    Organize & 47.3 & 36.0 & 1.3$\times$ \\
    \altrow
    Plug     & 97.6 & 70.0 & 1.4$\times$ \\
    Key      & 95.2 & 32.0 & 3.0$\times$ \\
    \altrow
    Tubes    & 95.2 & 92.0 & 1.0$\times$ \\
    \midrule
    Hardest five & 52.8 & 22.8 & 2.3$\times$ \\
    \altrow
    Insertion    & 96.0 & 64.7 & 1.5$\times$ \\
    All eight    & 69.0 & 38.5 & 1.8$\times$ \\
    \bottomrule
  \end{tabular}\end{adjustbox}
  \caption{\textbf{Ratio to the best leaderboard entry per task} (first-attempt success, \%). Ours: \method; Best: best entry; Ratio: Ours / Best. The largest ratio is 8.2$\times$ (Egg); the group rows compare the group averages; Pack and Mugs: mean of base and random layouts.}
  \label{supp:tab:gains}
\end{table}

\paragraph{Per-seed results.}
\Cref{supp:tab:seeds} lists the first-attempt successes of \method per official seed.

\begin{table}[t]
  \centering
  \footnotesize
  \setlength{\tabcolsep}{2pt}
  \tabstyle
  \begin{adjustbox}{max width=\linewidth}\begin{tabular}{@{}lcccc@{}}
    \toprule
    \headrow
    \hd{Task} & \hd{Seed 0} & \hd{Seed 1} & \hd{Seed 2} & \hd{Total (\%)} \\
    \midrule
    \texttt{fill\_egg\_holder}        & 31/64 & 34/63 & 27/61 & 92/188 (48.9) \\
    \altrow
    \texttt{store\_tools\_in\_toolbox} & 42/55 & 48/55 & 41/55 & 131/165 (79.4) \\
    \texttt{pack\_objects\_into\_boxes} & 31/55 & 30/55 & 35/55 & 96/165 (58.2) \\
    \altrow
    \begin{tabular}[c]{@{}l@{}}\texttt{pack\_objects\_into\_boxes}\\\texttt{\_random}\end{tabular} & 11/30 & 10/30 & 12/30 & 33/90 (36.7) \\
    \texttt{hang\_mugs}               & 20/30 & 13/30 & 16/30 & 49/90 (54.4) \\
    \altrow
    \texttt{hang\_mugs\_random}        & 6/30  & 8/30  & 11/30 & 25/90 (27.8) \\
    \texttt{organize\_table}          & 26/55 & 25/55 & 27/55 & 78/165 (47.3) \\
    \altrow
    \texttt{plug\_in\_charger}        & 54/55 & 53/55 & 54/55 & 161/165 (97.6) \\
    \texttt{insert\_key}              & 53/55 & 51/55 & 53/55 & 157/165 (95.2) \\
    \altrow
    \texttt{insert\_tubes}            & 52/55 & 53/55 & 52/55 & 157/165 (95.2) \\
    \bottomrule
  \end{tabular}\end{adjustbox}
  \caption{\textbf{First-attempt success of \method per official seed} (totals as in the main results; for \texttt{pack\_objects\_into\_boxes} and \texttt{hang\_mugs}, the base and random layouts are given separately, whose means are the main results, with 55 base and 30 random layouts per seed for \texttt{pack\_objects\_into\_boxes} and 30 each for \texttt{hang\_mugs}).}
  \label{supp:tab:seeds}
\end{table}

\paragraph{Random layouts of \texttt{pack\_objects\_into\_boxes}.}
The task has two layout configurations, base and random; the random layouts use object models that do not appear in the base layouts.
The harness handles random-layout failures with two modules, localizer training on closed-loop mid-episode frames and box re-alignment, whose settings are chosen on self-generated development layouts; the head-camera keypoint localizer is trained only on self-generated layouts with base object models and never sees a random-layout object model.
The harness succeeds on 96/165 (58.2\%) base and 33/90 (36.7\%) random layouts (95\% Wilson intervals 50.6--65.4 and 27.4--47.0); their mean, as computed by the leaderboard, is 47.4\% against 30.0\% for the best entry, PyRUA-Lean~\cite{si2026pyrualean}, whose 40\% on base and 20\% on random layouts lie below these intervals.
The two modules, added for the random layouts, also help on the base layouts: before them, the harness succeeds on 83/165 base layouts.
Most failures leave an object outside the box (shoe, electric toothbrush, car), followed by the orientation of the toothbrush in the box and, on random layouts, the facing direction of the car.

\section{Building the Tools}
\label{supp:build}

\subsection{Privileged Teachers}
\label{supp:teacher}

\paragraph{Replica and entry states.}
The physical parameters of each Isaac Lab~\cite{mittal2025isaaclab} replica (friction, inertia, solver iterations) are aligned with the native simulator.
The entry states of a stage are a bank of end states that the previous stage reaches in the native simulator, so upstream errors are part of training.

\paragraph{Reward.}
The reward is potential-based,
\begin{equation}
  r_t = \alpha\,(\Phi_t - \Phi_{t-1}) - c + r^{\text{term}}_t,
  \label{supp:eq:reward}
\end{equation}
where the stage potential $\Phi$ measures progress toward the stage goal (\eg, the remaining distance and angle to a grasp or handover pose), $c$ is a step cost, and $r^{\text{term}}_t$ rewards success and penalizes failure at the end of the stage.

\paragraph{Chunked TD3.}
The teacher outputs a chunk of $C = 5$ actions.
The twin critics evaluate a whole chunk, with the target
\begin{equation}
  y_t = \sum_{k=0}^{C-1} \gamma^k r_{t+k} + \gamma^{C} \min_{j=1,2} Q_{\phi'_j}\big(s_{t+C},\, \pi_{\theta'}(s_{t+C}) + \epsilon\big),
  \label{supp:eq:target}
\end{equation}
where $\phi'_j$ and $\theta'$ are the parameters of the target critics and the target actor, and $\epsilon$ is clipped Gaussian target-policy smoothing noise (standard deviation 0.2, clip 0.5).
As in TD3~\cite{fujimoto2018td3}, the actor is updated once every two critic updates, and the target networks are updated softly, $\phi'_j \leftarrow \tau\phi_j + (1-\tau)\phi'_j$ and $\theta' \leftarrow \tau\theta + (1-\tau)\theta'$, with $\tau = 0.005$.
The actor loss normalizes the critic term and regularizes the actor toward an anchor policy $\bar{\pi}$,
\begin{equation}
  \mathcal{L}_{\text{actor}}(\theta) = \mathbb{E}_{s}\Big[-\frac{Q_\phi\big(s, \pi_\theta(s)\big)}{\mathbb{E}\,\lvert Q_\phi \rvert} + \lambda\, \big\lVert \pi_\theta(s) - \bar{\pi}(s) \big\rVert^2 \Big].
  \label{supp:eq:actor}
\end{equation}

\paragraph{Phase 1: reference anchor.}
Early in training, $\bar{\pi}$ is a rule-based reference controller evaluated online at the visited states, unlike offline behavior-cloning regularization~\cite{fujimoto2021td3bc}, which targets the actions of a fixed dataset.
The replay buffer is first filled with 4 rounds of rollouts from the reference controller with exploration noise (standard deviation 0.1); the actor is then pretrained by behavior cloning on the reference for 2000 steps, and for the first 20 iterations only the critic is trained, so that it is fitted on transitions that already reach the goal before the actor relies on it.
Afterwards, $\lambda$ decays linearly from 50 to 5.
During this phase the actor also receives the reference chunk as input, dropped as a whole with probability 0.5; the teacher is evaluated and deployed with the reference input set to zero.

\paragraph{Phase 2: best-policy anchor.}
When the evaluated success of the teacher improves by less than 0.003 over 2 or 4 consecutive evaluations (and at least 40 iterations have passed), the reference is removed.
The actor is reloaded from the best policy found so far, a new Adam optimizer with learning rate $3\times10^{-5}$ is created, and the critic alone is retrained for 5 iterations.
The anchor $\bar{\pi}$ is then the best policy found so far and is refreshed whenever a new best is found; its weight is $\lambda = 30$ in most runs.
On \texttt{plug\_in\_charger}, the grasping, carrying, and handover teachers passed the gate in phase~1.

\paragraph{Networks and optimization.}
The actor and the twin critics are MLPs with three hidden layers of 256 units, layer normalization, and GELU activations; the actor outputs a $\tanh$-bounded chunk.
Each iteration collects one batch of episodes with the current actor (exploration noise 0.1) and then performs $8\lceil T/C \rceil$ updates for a stage horizon of $T$ steps (update-to-data ratio 8), with batch size 4096, a replay buffer of $2\times10^{6}$ transitions, Adam with learning rate $3\times10^{-4}$, and gradient clipping at 10.
On \texttt{plug\_in\_charger}, a teacher is trained for 300 iterations with 512 environments.
The checkpoint with the highest evaluated success is kept.
A teacher is accepted only when its per-stage and full-chain success on held-out layouts in the native simulator passes the gate $\gamma$, both with and without perturbations of the entry state.
\Cref{supp:tab:td3} summarizes the differences from TD3.

\begin{table}[t]
  \centering
  \footnotesize
  \setlength{\tabcolsep}{3pt}
  \tabstyle
  \begin{adjustbox}{max width=\linewidth}\begin{tabular}{lll}
    \toprule
    \headrow
    \hd{Item} & \hd{TD3} & \hd{Ours} \\
    \midrule
    Setting & online, off-policy & online, off-policy, parallel replica \\
    \altrow
    Action & single step & chunk of $C=5$; critic on chunk \\
    Critics & twin, smoothing, delay & same (0.2 / 0.5, delay 2, $\tau = 0.005$) \\
    \altrow
    Actor loss & $-Q$ & $-Q/\mathbb{E}\lvert Q\rvert + \lambda \lVert \pi - \bar{\pi} \rVert^2$ \\
    Anchor $\bar{\pi}$ & -- & reference, then best policy \\
    \altrow
    Initialization & random & BC on reference, 20 critic-only iters \\
    \bottomrule
  \end{tabular}\end{adjustbox}
  \caption{\textbf{Teacher training compared with TD3}~\cite{fujimoto2018td3}. The critics and target updates follow TD3. The teacher acts in action chunks, anchors the actor to a reference controller and later to its best policy, and is initialized by behavior cloning on the reference (\Cref{supp:teacher}).}
  \label{supp:tab:td3}
\end{table}

\subsection{Student Distillation}
\label{supp:student}

All accepted teachers are distilled into one $\pi_{0.5}$ student $\hat{\pi}$ conditioned on the stage instruction, which sees only the three camera images and proprioception $o_t$, whereas the teacher $\pi^\star$ sees the simulator state $s_t$.
Data are collected in the native simulator and aggregated over iterations $k = 0, \dots, K$ following DAgger~\cite{ross2011dagger}:
\begin{equation}
  \mathcal{D}_{k+1} = \mathcal{D}_{k} \cup \big\{ \big(o_t,\, \pi^\star(s_t)\big) \big\}_{t \sim \beta_k \pi^\star + (1-\beta_k)\hat{\pi}_k}.
  \label{supp:eq:dagger}
\end{equation}
In iteration $k$, each action chunk is executed by the teacher with probability $\beta_k$ and by the current student $\hat{\pi}_k$ otherwise, so the two policies are mixed per chunk rather than per step as in the original formulation, and every visited state is labeled with the teacher's clean action chunk; $\beta_k$ decreases from 1 to 0 over the iterations.
To cover states off the teacher's path, iteration~0 also executes teacher chunks with DART-style noise~\cite{laskey2017dart} in half of the episodes, while the labels remain the teacher's clean actions at the states actually reached, and perturbs the stage entries; failed episodes are kept.
Segments in which the robot is stuck are kept with a reduced weight, so that these erroneous segments count less in training.
After each iteration, $\hat{\pi}_{k+1}$ is retrained on $\mathcal{D}_{k+1}$ from the $\pi_{0.5}$ base checkpoint rather than from $\hat{\pi}_k$.
\Cref{supp:tab:dagger} summarizes the differences from DAgger.

The student outputs a chunk of actions (joint increments and absolute gripper commands) and executes only its first actions before the next inference.
Only the action expert is trained, with a flow-matching loss, while the vision tower and the language model stay frozen; no low-rank adaptation is used.
\Cref{supp:tab:studenthp} lists the values used for data collection and training.

\paragraph{DAgger and data coverage make the student reliable.}
With behavior cloning on 451 episodes, the handover student of \texttt{plug\_in\_charger} reaches 22/25, 25/28 after DAgger, and 28/28 (197/200 on unseen layouts) after full-range layouts and further DAgger rounds; with too little data (316 episodes), it releases the charger early and carry success drops to about 50\%.

\paragraph{Why the vision encoder stays frozen.}
Two alignment students are trained from the takeover-probe student on identical data with the same seed, once with the encoder frozen and once unfrozen (settings in \Cref{supp:tab:studenthp}): after 3k steps they reach 2/28 (frozen) and 3/28 (unfrozen) in the takeover probe, and both 0/28 after 1k and 2k steps, so unfreezing gives no measurable gain.
On \texttt{insert\_key}, unfreezing even hurts: one DAgger round of the grasp--lift and handover student, trained twice in the same way, completes the handover on 90 (frozen) and 35 (unfrozen) of 100 paired layouts from the training distribution, and on 88 and 60 with a second weighting of the DAgger data.
The deployed students therefore keep the vision encoder frozen.

\paragraph{Takeover probe on \texttt{insert\_key}.}
The takeover probe, repeated on the insertion stage of \texttt{insert\_key}, gives the same split: a $\pi_{0.5}$ student distilled from 1471 teacher demonstrations of this stage with the same recipe succeeds on 3/32 held-out layouts, against 26/28 for the teacher.

\begin{table}[t]
  \centering
  \footnotesize
  \setlength{\tabcolsep}{3pt}
  \tabstyle
  \begin{adjustbox}{max width=\linewidth}\begin{tabular}{lll}
    \toprule
    \headrow
    \hd{Item} & \hd{DAgger} & \hd{Ours} \\
    \midrule
    Mixing & per step & per 5-step chunk \\
    \altrow
    $\beta$ schedule & $\beta_1 = 1$, then decay & 1, 0.5, 0.3, 0.1, 0 \\
    Labels & $\pi^\star(s)$ on visited states & teacher chunk on visited states \\
    \altrow
    Retraining & on aggregated $\mathcal{D}$ & on aggregated $\mathcal{D}$, from base \\
    Observations & shared & teacher: state; student: RGB \\
    \bottomrule
  \end{tabular}\end{adjustbox}
  \caption{\textbf{Student data collection compared with DAgger}~\cite{ross2011dagger}. As in DAgger, the student is retrained on teacher-labeled data aggregated over iterations. Each action chunk, rather than each step, is executed by the teacher with probability $\beta$ and otherwise by the student. Each retraining starts from the $\pi_{0.5}$ base checkpoint, and only the teacher sees the simulator state (\Cref{supp:student}).}
  \label{supp:tab:dagger}
\end{table}

\begin{table}[t]
  \centering
  \footnotesize
  \setlength{\tabcolsep}{4pt}
  \tabstyle
  \begin{adjustbox}{max width=\linewidth}\begin{tabular}{ll}
    \toprule
    \headrow
    \hd{Item} & \hd{Value} \\
    \midrule
    Action chunk / executed per inference & 50 / 5 \\
    \altrow
    $\beta_k$, $k = 0, \dots, 4$ & 1, 0.5, 0.3, 0.1, 0 \\
    DART noise (iteration~0), std. & 0.15 \\
    \altrow
    Stage-entry perturbation & ${\approx}$1\,cm, 5$^\circ$ (max.\ 3\,cm, 15$^\circ$) \\
    Weight of stuck segments & 0.2 \\
    \altrow
    Trained: action expert & ${\sim}$430\,M parameters \\
    Frozen: vision tower, language model & ${\sim}$2.9\,B parameters \\
    \altrow
    GPUs & 4 $\times$ H200 \\
    Batch size & 640 \\
    \altrow
    Learning rate & $8\times10^{-5}$, cosine decay \\
    Steps per round & 2.5--3.3k \\
    \altrow
    Encoder ablation (lr / batch / steps) & $5\times10^{-5}$ / 96 / 3k \\
    \bottomrule
  \end{tabular}\end{adjustbox}
  \caption{\textbf{Student hyperparameters} for data collection and training (\Cref{supp:student}); the last row is the encoder ablation.}
  \label{supp:tab:studenthp}
\end{table}

\subsection{Keypoint Localizer}
\label{supp:localizer}

\paragraph{KPNet.}
Keypoints are defined on the object's CAD model, and their labels are the projections of these points under ground-truth poses; training images come from teacher and student rollouts and therefore include gripper occlusion.
KPNet is a ResNet-18~\cite{he2015resnet} encoder (ImageNet initialization) with an FPN-style decoder (128-channel lateral connections).
It takes the full $640\times480$ image and predicts one stride-4 heatmap per keypoint, converted to full-resolution pixel coordinates by soft-argmax, and one visibility logit per keypoint from pooled encoder features; it has 11.45--11.46\,M parameters depending on the number of keypoints.
The loss is an L1 loss on visible keypoints plus a binary cross-entropy loss on visibility, optimized with AdamW (learning rate $3\times10^{-4}$ for the head and $10^{-4}$ for the backbone, one warm-up epoch, cosine decay); the epoch with the best development error is kept.
The pose is recovered by PnP on the CAD keypoints; objects on the table use a plane-constrained fit.
As a self-check, the measured offset must be near zero at the end of successful teacher insertions.

\paragraph{Camera choice.}
The localizer of \texttt{plug\_in\_charger} uses the wrist cameras, chosen before training by a sensitivity check: the agent adds pixel noise to ground-truth keypoints and recovers the plug-to-outlet offset by PnP.
On the wrist camera, 3\,px of noise gives a median offset error of 0.9\,mm (94\% below 2\,mm), whereas on the head camera 1\,px already corresponds to about 15\,mm.

\paragraph{Other tasks.}
On \texttt{fill\_egg\_holder}, a wrist KPNet predicts the egg center and both ends, and the egg pose is fitted with the known half-length of 28.8\,mm: the median center error is 0.51\,mm (p90 2.08\,mm, 88.8\% below 3\,mm).
A head-camera KPNet locates the holder and the basket; on held-out official layouts of seed~0 (62 frames), the holder error is 0.26\,mm median (0.71\,mm maximum) and the basket error 1.96\,mm median.
On \texttt{store\_tools\_in\_toolbox}, SegNet (ResNet-34 with an FPN decoder, 10 classes, 16.3k images, 12 epochs) reaches a held-out tool-mask mIoU of 0.954 and a slot mIoU of 0.853, against a tool-mask mIoU of 0.594 for SAM3~\cite{carion2026sam3} zero-shot; a KPNet additionally measures the pose of the tool in the gripper.

\subsection{Tool Ablations}
\label{supp:toolabl}

The tool ablations of \Cref{tab:ablation} run on all 55 official layouts of seed~0; every variant removes tools from the final inner harness of the task.
Without VLA skills, the VLM executes their stages with the base motion tool; without visual localizers, it aligns from the RGB images; without scripts, it executes the scripted stages chunk by chunk, with at most 5 control steps per chunk and the same localizer readings as the scripts.
On \texttt{insert\_key}, only two ablated episodes, both without visual localizers, insert the key; most of the others lift the key but do not insert it.
The native criterion requires the key within 7\,mm of the slot, tilted by at most 7$^\circ$, and turned by at least 30$^\circ$ at the same time.
Without localizers, this offset must be judged from the RGB images.
Without VLA skills, the carry and insertion scripts still estimate the key pose in the hand from the nominal grasp of the skill, so a different grasp by the VLM shifts their estimate.
Without scripts, the VLM receives the same KPNet measurement but corrects the key once per chunk instead of at every control step, and every episode reaches the 300-step limit, whereas the full harness needs a median of 222 steps.
On \texttt{insert\_tubes}, without scripts the VLM executes the insertion itself, and success is 85.5\% (exact McNemar test against the full harness on the same layouts, $p{=}0.29$) within the 500-step limit: a tube counts as inserted when it is in the slot with a tilt of up to 30$^\circ$, and no turn is required.
Without localizers, the scripts of \texttt{insert\_tubes} lose their input, and the VLM controls the whole episode.
Overall, the VLM handles stages with a coarse tolerance, such as grasping and lifting, but stages that need millimeter alignment fail without the tool that measures or controls at that precision.

\section{Inner Harness Interface}
\label{supp:interface}

Three rules keep the VLM the judge of progress.
First, skills return observations rather than judgments of success, namely completion, steps used, measurements, and new images.
Second, a skill card states the skill's function, precondition, cost in control steps, and measured reliability, but no task plan.
Third, a skill whose precondition is violated returns without acting, \eg, when nothing is held or the target is out of view.

The router $\mathcal{R}$, written by the outer agent, lists for each task the available skills, their preconditions, and their measured reliability; the VLM reads it and, at each stage, selects a skill from it.

\paragraph{Self-revision.}
Self-revision changes only the task notes, as in experiential-learning agents~\cite{zhao2024expel}.
The VLM starts from the tools and their skill cards without any task note and makes one first attempt on each self-generated layout; after each episode, it receives the judged result, revises its task notes, and writes a summary, and the next episode starts on a new layout with the revised notes.
After the last episode, the notes are frozen, and the harness is tested on the official layouts.
On \texttt{plug\_in\_charger}, after 20 episodes, the frozen notes reach 97.6\% on all official layouts.

\section{Outer Harness}
\label{supp:outer}

\begin{algorithm}[t]
  \caption{Outer-harness loop of \method}
  \label{supp:alg:outer}
  \small
  \algrenewcommand\algorithmicrequire{\textbf{Input:}}
  \algrenewcommand\algorithmicensure{\textbf{Output:}}
  \begin{algorithmic}[1]
    \Require task $\mathcal{T}$; VLM $\mathcal{M}$ with base tools $\mathcal{K}_0$
    \Ensure tools $\mathcal{K}$ and router $\mathcal{R}$
    \State split $\mathcal{T}$ into stages $S_1, \dots, S_n$ where the failures of $(\mathcal{M}, \mathcal{K}_0)$ cluster, with official demonstrations as reference where available
    \State train a privileged teacher per $S_i$; distill a VLA student
    \State give each $S_i$ the VLA, a visual localizer, or $\mathcal{M}$ by the takeover probe; write $\mathcal{R}$
    \Repeat
      \State roll out $(\mathcal{M}, \mathcal{R}, \mathcal{K})$; find the first failing stage, $i^\star \gets \min\{i : \hat{p}_i < \gamma\}$
      \State in parallel, improve the tool of $S_{i^\star}$ on its failures and train other tools for $S_{i^\star}$
      \State give $S_{i^\star}$ the best of them (takeover probe) if $\hat{p}_{i^\star}$ rises; update $\mathcal{R}$; log it in $\mathcal{E}$
    \Until{$\hat{p}_i \geq \gamma$ for all $i$, or the budget is spent}
  \end{algorithmic}
\end{algorithm}

\begin{figure*}[t]
  \centering
  \includegraphics[width=\linewidth]{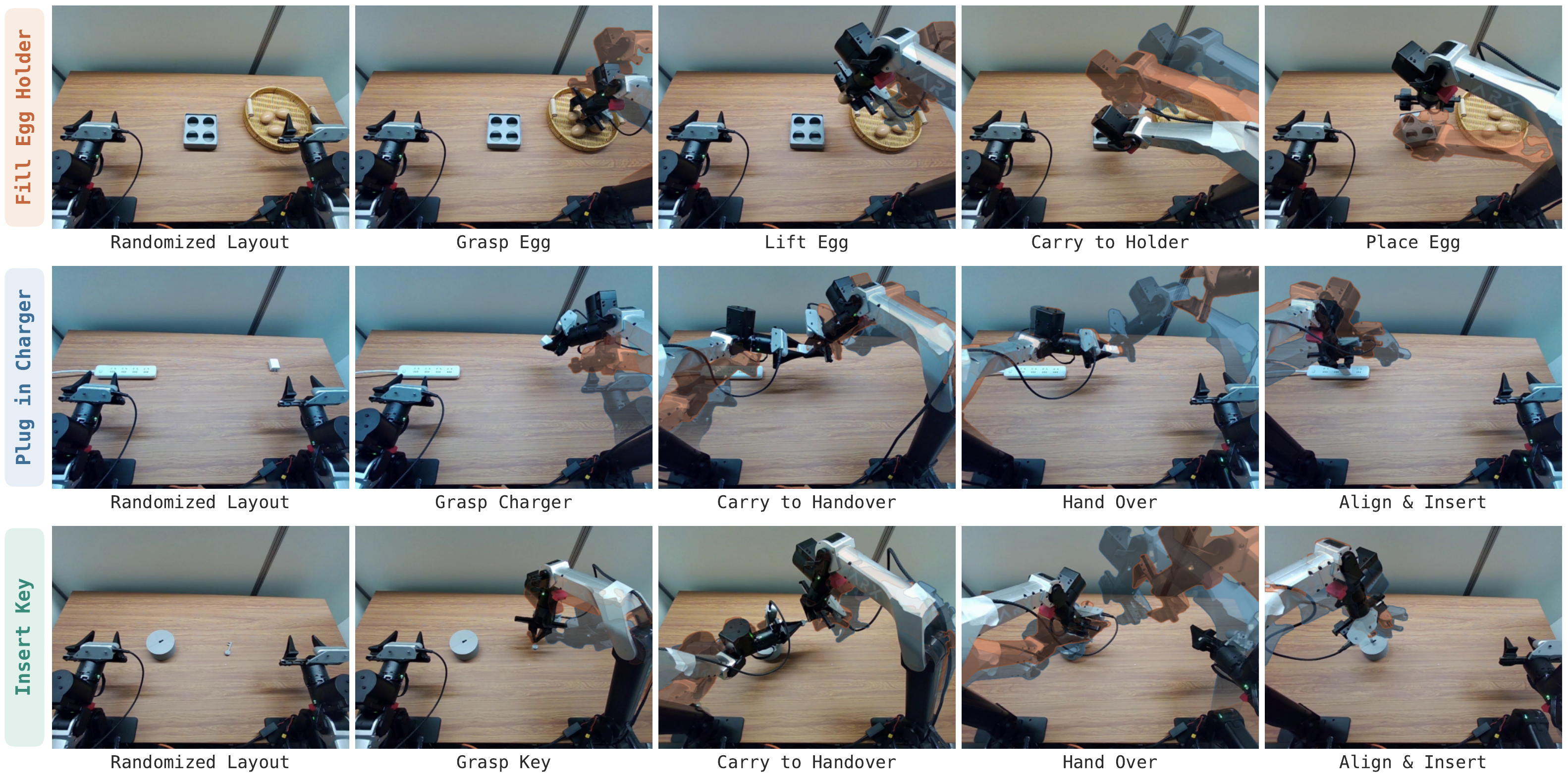}
  \caption{\textbf{Stage decomposition} found by the outer agent in simulation, shown on the real robot. Each row starts from a randomized layout and shows the end of each stage: \texttt{fill\_egg\_holder} (grasp, lift, carry, place), \texttt{plug\_in\_charger} and \texttt{insert\_key} (grasp, carry to the handover pose, hand over, align and insert). For \texttt{plug\_in\_charger}, the last column shows the end of alignment and insertion (S4--S5); the final release and arm reset (S6) are not shown.}
  \label{supp:fig:stages}
\end{figure*}

\begin{figure}[t]
  \centering
  \includegraphics[width=\linewidth]{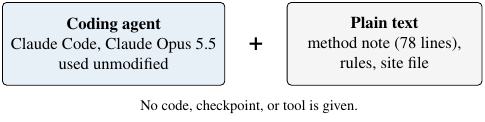}
  \caption{\textbf{The complete outer harness}: an unmodified coding agent and the plain text it reads. The method note states the three decisions (decompose, allocate, revise) in a few sentences and gives a generic training recipe; every task-specific decision is made by the agent. The builds of \Cref{supp:tab:builds} start from the same method note.}
  \label{supp:fig:outer}
\end{figure}

\noindent \Cref{supp:alg:outer} summarizes the loop of the outer harness, and \Cref{supp:fig:outer} shows the complete outer harness that runs it: an unmodified coding agent and a method note that states each of the three decisions in a few sentences and gives a generic training recipe for the teachers, students, and localizers.
The stages are found in simulation; \Cref{supp:fig:stages} shows them on the real robot for the three tasks deployed there, where the same stages are used unchanged.

\paragraph{A minimal outer harness.}
The outer agent is an off-the-shelf coding agent used without modification, and the outer harness has no component of its own beyond the files of \Cref{supp:fig:outer}: a method note of 78 lines with the three decisions and recipes for the replica, teacher, student, localizer, and harness, 23 lines of rules, a site file with machines and accounts, and a watchdog that only restarts the session.
The recipes are generic and name no task or object; the agent makes every task-specific decision, from the stage boundaries, rewards, reference scripts, and training replica to the tool type of each stage, the recorded data, and each revision.
Every component of the inner harness, including the code of the takeover probe and of the revisions, is written or trained by the agent.
The decisions are sentences of advice that the agent carries out, so the outer harness has no module whose removal would be a meaningful ablation; instead, the alternatives to each decision are measured (\Cref{supp:tab:outeralt}).

\paragraph{Experience library.}
Besides stage recipes, acceptance protocols, failure--fix pairs, and checkpoints, the library $\mathcal{E}$ holds engineering lessons, such as slimming simulation assets, scheduling GPU jobs, and keeping the training replica consistent with the native simulator.

\paragraph{Building from scratch.}
With the outer harness of \Cref{supp:fig:outer}, the outer agent builds an inner harness from scratch: it starts with an empty experience library, without code or checkpoints of earlier builds, with the method note, and with a 72\,h budget.
It develops only on layouts that it samples from the task configuration; each frozen version is tested once on official layouts, and the result is not returned to the agent, which selects versions on self-generated held-out layouts.
On \texttt{plug\_in\_charger}, the first frozen version already reaches 94.5\% on official seed~1 after 7.1\,h, with one human intervention: a correction to write the stage decomposition before recording student data.
The main gains came from the loop itself: a takeover probe showed that $\pi_{0.5}$ cannot align or insert (0/20 each), and a fix for an IK-branch stall before the handover, found among the development failures, raised development success from 85.5\% to 94.5\%.
The other builds run without human intervention: on official seed~1, \texttt{insert\_key} reaches 90.9\% after 4.5\,h and 94.5\% after 8.0\,h, \texttt{fill\_egg\_holder} reaches 46.0\% after 3.0\,h and 47.6\% after 12.1\,h (29/63 and 30/63; as in the main evaluation, the two layouts of seed~1 that fail at reset are discarded), and \texttt{hang\_mugs} reaches 52.0\% after 12.8\,h (32.0\% on \texttt{hang\_mugs\_random}), tested on the first 25 layouts of seed~1 (13/25 and 8/25).
\Cref{supp:tab:keybuild} lists every tested version of Key.
Before its first tested version, moving the grasp toward the center of mass of the key stopped drops at the handover; afterwards, a learned key-tip localizer raised development success from 56\% to 75\%.
From v5 on, official success stays at 91--95\%, and the outer agent adds tools for perturbed execution: on self-generated layouts with a 2\,mm calibration error, success rises from about 20\% to 91\%.

\paragraph{Alternatives to the outer agent's decisions.}
Deleting passages of the method note would not isolate a component, since each build is a single multi-hour run of a stochastic agent.
\Cref{supp:tab:outeralt} collects them: without a build, the same VLM succeeds on no task; assigning a stage to another tool type than the one chosen by the takeover probe loses most of the success; and the builds from scratch reach 94.5\% and 90.9\% starting from an empty experience library.

\begin{table}[t]
  \centering
  \footnotesize
  \setlength{\tabcolsep}{3pt}
  \tabstyle
  \begin{adjustbox}{max width=\linewidth}\begin{tabular}{>{\raggedright\arraybackslash}p{0.26\linewidth} >{\raggedright\arraybackslash}p{0.40\linewidth} >{\raggedright\arraybackslash}p{0.25\linewidth}}
    \toprule
    \headrow
    \hd{Decision} & \hd{Simpler alternative} & \hd{SR (\%)} \\
    \midrule
    Build the inner harness & none: VLM with base tools $\mathcal{K}_0$ & 0.0, all tasks \\
    \altrow
    Allocation by takeover probe & VLA skill on alignment & 0/28 \\
     & VLM on alignment (w/o localizers) & 20.0 / 3.6 / 0.0 \\
    \altrow
     & VLM on control stages (w/o VLA skills) & 9.1 / 0.0 / -- \\
     & VLM on scripted stages (w/o scripts) & -- / 0.0 / 85.5 \\
    \altrow
    Experience library & empty library & 94.5 / 90.9 / -- \\
    Revise loop & first tested version & 50.9 $\rightarrow$ 94.5 (Key) \\
    \bottomrule
  \end{tabular}\end{adjustbox}
  \caption{\textbf{Simpler alternatives to the decisions of the outer agent}, each measured as an inner-harness variant. Triples: Plug / Key / Tubes.}
  \label{supp:tab:outeralt}
\end{table}

\begin{table}[t]
  \centering
  \footnotesize
  \setlength{\tabcolsep}{3pt}
  \tabstyle
  \begin{adjustbox}{max width=\linewidth}\begin{tabular}{lccp{0.56\columnwidth}}
    \toprule
    \headrow
    \hd{Ver.} & \hd{Time} & \hd{SR} & \hd{Main change} \\
    \midrule
    v1 & 2.1\,h & 50.9 & classical localizers, stages chained by scripts \\
    \altrow
    v2 & 2.8\,h & 69.1 & VLM sequences the tools; learned key-tip localizer \\
    v3 & 3.6\,h & 67.3 & regrasp before insertion; second servo near the lock \\
    \altrow
    v4 & 4.5\,h & 90.9 & lower lift after the grasp \\
    v5 & 7.3\,h & 92.7 & visual closed loop at the handover; slip detection and re-insertion \\
    \altrow
    v6 & 7.7\,h & 92.7 & key-tip localizer v5; larger servo correction \\
    v7 & 8.0\,h & 94.5 & slow press until the key is seated; hold after the turn \\
    \altrow
    v8 & 8.4\,h & 90.9 & back off and re-press when jammed in the lock \\
    v9 & 9.5\,h & 94.5 & key-ring localizer (wrist) to detect slip during insertion \\
    \bottomrule
  \end{tabular}\end{adjustbox}
  \caption{\textbf{Versions of the \texttt{insert\_key} build from scratch.} Time: as in \Cref{supp:tab:builds}; SR: success on official seed~1 (\%), each version tested once. v4 is also tested once on official seed~2 (96.4\%).}
  \label{supp:tab:keybuild}
\end{table}

\section{Localizer Comparison}
\label{supp:loccmp}

The test frames are all calls of the localizer made by the VLM after the handover in the full runs on official seeds 0 and 2 (1174 frames, 963 of them with the gripper tip within 6\,cm of the power strip); every method receives the same three RGB images and end-effector poses.
Each method estimates the power-strip pose, from which the three outlet positions follow with the same CAD model; the error is their mean $xy$ distance to the simulator ground truth, and a frame without an estimate counts as a miss.
PyRUA-Lean~\cite{si2026pyrualean} is run unchanged (\texttt{locate}, SAM3 segmentation and multi-view geometry, RGB only).
For Auriga~\cite{wang2026auriga}, we run its object-table geometry on SAM3~\cite{carion2026sam3} masks and, in its favor, replace the VGGT depth with ground-truth depth. With SAM3.1 (facebook/sam3.1), which Auriga itself uses, on the 1212 new frames of \Cref{supp:tab:newframes}, the paired estimates differ from those with SAM3 by 0.02\,mm (head camera, median; 0.11\,mm p90) and 0.17\,mm (wrist camera; 0.44\,mm p90), and the fitted strip length is 14.2\,cm with both.

Per seed, KPNet is within 3\,mm on 70.2\% (seed~0) and 64.1\% (seed~2) of all frames, and PyRUA-Lean (head) on 49.8\% and 45.3\%.
A localizer call takes 0.13\,s for KPNet, 2--3\,s for PyRUA-Lean, and 0.3\,s for Auriga without VGGT.

\paragraph{New frames on 110 layouts.}
To score every method, including SAM3.1 and the ground-truth mask, on the same frames, we rerun the deployed harness on all 110 official layouts of seeds 0 and 2 (107/110 successes) and record all 1212 localizer calls after the handover (996 within 6\,cm of the power strip).

\paragraph{Closed-loop replacement.}
To compare the localizers on the states that each reaches itself, we replace the power-strip pose measured by KPNet in the deployed harness with the pose from Auriga (SAM3 masks and ground-truth depth, as above) and leave everything else unchanged.
The 52/55 of the deployed harness on seed~0 is from the same rerun.

\paragraph{Comparison with DINOv3.}
\Cref{supp:tab:dino} compares KPNet with keypoint heads on DINOv3~\cite{simeoni2025dinov3} features on a separate test set of \texttt{plug\_in\_charger}: frames of official opposite-side layouts in which the receiving gripper holds the charger with a true offset below 20\,mm (about 580 frames).
Both are trained on the same 85{,}815 frames; KPNet is fine-tuned end to end (12 epochs, about 1.7\,h on one GPU), whereas the DINOv3 heads are trained on frozen features under a 30-minute budget (2--5 epochs), so the comparison favors KPNet in training passes; zero-shot DINOv3 matching uses 16 labeled reference frames.
The DINOv3 heads of \Cref{supp:tab:dino} are trained on the same simulator labels as KPNet; that they also reach sub-millimeter error, whereas zero-shot matching with the same DINOv3 backbones errs by 20--35\,mm, shows that the gain comes from training keypoint localizers on simulator labels, which the outer agent generates; among such localizers, KPNet is the more accurate and the smaller (11.5\,M parameters).

\begin{table}[t]
  \centering
  \footnotesize
  \setlength{\tabcolsep}{3pt}
  \tabstyle
  \begin{adjustbox}{max width=\linewidth}\begin{tabular}{lccc}
    \toprule
    \headrow
    \hd{Method} & \hd{Params} & \hd{Med.\ / p90 (mm)} & \hd{${<}3$\,mm} \\
    \midrule
    DINOv3 ViT-B, zero-shot match & 86\,M & 35.1 / 50.7 & 0.5\% \\
    \altrow
    DINOv3 ViT-L, zero-shot match & 303\,M & 20.7 / 38.0 & 1.4\% \\
    DINOv3 ViT-L, frozen + head & 303\,M + 2.8\,M & 0.61 / 1.35 & 96.2\% \\
    \altrow
    DINOv3 ViT-B, frozen + head & 86\,M + 2.2\,M & 0.44 / 1.03 & 97.0\% \\
    \midrule
    \oursrow
    KPNet (ours) & 11.5\,M & \textbf{0.13} / \textbf{0.40} & \textbf{98.5\%} \\
    \bottomrule
  \end{tabular}\end{adjustbox}
  \caption{\textbf{KPNet vs.\ DINOv3 keypoint localization} on \texttt{plug\_in\_charger}. Error of the measured offset that alignment corrects, on wrist-camera frames in which the receiving gripper holds the charger within 20\,mm of the target. The trained methods use the same simulator labels (\Cref{supp:localizer}).}
  \label{supp:tab:dino}
\end{table}

\section{Deployment Efficiency}
\label{supp:efficiency}

\paragraph{Real robot.}
Direct VLM control is not run on the real robot: it does not complete these tasks in simulation, and a real trial would keep the robot under VLM control for 18--54\,min at about 25\,s per call (24\,s median in simulation, below).

\paragraph{Simulation.}
Under direct control, every 5 control steps cost one VLM call: on \texttt{plug\_in\_charger}, a call takes 24\,s (median), and an episode reaches the 400-step limit after 80 calls and 33\,min without succeeding; direct control spends 99\% of its time in the VLM.
The efficiency comparison in simulation counts Astra-issued action chunks plus tool calls as VLM calls, measures the time to success capped at the episode limit (a failed episode counts as the cap), and reports medians over 15 direct-control episodes and 165 episodes of \method; the simulator does not advance while the VLM thinks.
These times include the VLM thinking time.

\section{Real-Robot Experiments}
\label{supp:real}

Videos of the real-robot trials of both methods and of the corresponding simulated episodes are on the project page.

\section{Additional Related Work}
\label{supp:related}

Several recent systems also let coding agents build or revise the tools around a robot policy.
MAESTRO~\cite{shi2025maestro} writes robot programs over a curated library of perception, planning, and control modules, including VLAs; URAI~\cite{ge2026urai} couples a programming agent that constructs robot tools with an execution agent that calls them; CaP-X~\cite{fu2026capx} benchmarks and improves coding agents for manipulation; Meta-Harness~\cite{lee2026metaharness} searches over harness code using the traces of earlier candidates; and, concurrently, RoboArchitect~\cite{wu2026roboarchitect} rewrites the control code around a frozen robot policy.
In these systems the tools are code or given modules; \method additionally trains learned tools in simulation, VLA skills and keypoint localizers, and decides per stage which type of tool is needed.

\end{document}